\documentclass{article}

\PassOptionsToPackage{numbers, compress}{natbib}
\usepackage[preprint]{neurips_2026}

\usepackage[utf8]{inputenc} % allow utf-8 input
\usepackage[T1]{fontenc}    % use 8-bit T1 fonts
\usepackage{hyperref}       % hyperlinks
\usepackage{url}            % simple URL typesetting
\usepackage{fontawesome5}   % icons (e.g. project page globe icon)
\usepackage{booktabs}       % professional-quality tables
\usepackage{amsfonts}       % blackboard math symbols
\usepackage{nicefrac}       % compact symbols for 1/2, etc.
\usepackage{microtype}      % microtypography
\usepackage{subcaption}
\usepackage{graphicx}
\usepackage{float}
\usepackage{algorithm}
\usepackage{algpseudocode}
\usepackage{array}        % for p{} columns
\usepackage{ragged2e}     % for \RaggedRight
\usepackage{makecell}     % optional: for nicer multiline cells
\usepackage{multirow}

\usepackage{amsmath}

\DeclareMathOperator*{\argmin}{arg\,min}

\usepackage{siunitx}
\usepackage[dvipsnames]{xcolor}
\usepackage{tikz}
\usetikzlibrary{calc}

\newcommand{\methodname}{Robust Multi-Stage Implicit Maximum Likelihood Estimation}
\newcommand{\methodshort}{ROMS-IMLE}

\title{ROMS-IMLE: A Minimalist Approach to Competitive Single-Step Generative Modelling}

\author{%
  Chirag Vashist \\
  APEX Lab \\
  Simon Fraser University\\
  \texttt{chirag\_vashist@sfu.ca} \\
  \And
  Ke Li \\
  APEX Lab \\
  Simon Fraser University, Amii and CIFAR \\
  \texttt{keli@sfu.ca} \\
}

\begin{document}

% Prevent \flushbottom (set by neurips_2026.sty) from stretching vertical
% glue around section headings to justify short pages; let slack fall to the
% page bottom instead.
\raggedbottom

\maketitle

\definecolor{projectlinkcolor}{HTML}{553FC3}
\begin{center}
    \textcolor{projectlinkcolor}{\href{https://serchirag.github.io/roms-imle/}{\faGlobe~\url{https://serchirag.github.io/roms-imle/}}}
\end{center}

\begin{abstract}
    Generative models have undergone many generations of evolution, from VAEs/GANs to diffusion/flow matching. Along the way, the underlying techniques have become more complicated and various beliefs about what drives strong empirical performance have taken hold. Due to the success of diffusion models and flow matching, one of the more common beliefs is the importance of transforming the noise distribution to the data distribution gradually through many small transformations. 
     We ask whether this is truly necessary, and take a minimalist approach to designing a competitive generative model. We start with the bare-bones essentials, namely just a training objective and a model. We purposefully make both simple. For the training objective, we choose Implicit Maximum Likelihood Estimation (IMLE), and eschew more complicated alternatives such as variational inference, adversarial training and numerical integration. For the model, we eschew transformers and instead choose a moderately sized convolutional network. Then we judiciously added elements that are truly essential, which surprisingly do \emph{not} include iterative denoising. The result is a single-step parameter-efficient generative model that produces high quality samples at fast speed: it achieves an FID of 2.56 on ImageNet 256 and simultaneously attains good precision and recall. 
\end{abstract}

\section{Introduction}

Image generation has advanced through a succession of paradigms over the past decade. Variational autoencoders (VAEs)~\cite{kingma2022autoencodingvariationalbayes} and generative adversarial networks (GANs)~\cite{goodfellow2014generativeadversarialnetworks} were among the first models to generate realistic images in a single forward pass, yet each approach carries a characteristic weakness. GANs are unstable to train and prone to mode collapse, which limits the diversity of their samples. VAEs train stably but tend to produce blurry, low-fidelity samples. Stochastic interpolant models (SI for short), including diffusion and flow-matching models~\cite{ho2020denoisingdiffusionprobabilisticmodels, song2021scorebasedgenerativemodelingstochastic, lipman2023flowmatchinggenerativemodeling, stoc-inter}, resolved many of these issues and now define the state of the art in image synthesis. Inspired by this success, subsequent work has grown steadily more complicated, introducing elaborate architectures~\cite{peebles2023scalable, ma2024sit}, sophisticated sampling schemes~\cite{edm, song2022denoisingdiffusionimplicitmodels}, and intricate training objectives~\cite{song2023consistencymodels, meanflow, zhou2025inductive, frans2024stepdiffusionshortcutmodels}. 

We ask whether all of the complex machinery introduced in recent years is truly necessary. Rather than adding to it, we take a minimalist approach: we start from the simplest generative model we can and add only the components from modern approaches that prove essential. We choose to build on Implicit Maximum Likelihood Estimation (IMLE)~\cite{li2018implicitmaximumlikelihoodestimation}, which offers a stable and mode-covering training objective that is simple and intuitive. IMLE works by matching each data point to its nearest sample from the model, and then training the model to reduce the distance between them. IMLE optimizes a likelihood-based objective akin to regression, without the variational bounds and encoders of VAEs, the adversarial training of GANs, or the iterative numerical integration used by stochastic interpolant models.  

The advantage of modern SI models over earlier single-step models is commonly attributed to how they decompose generation~\cite{trilemma, diffusion-beats-gan}: instead of mapping the prior to the data distribution in one large transformation, they break it into a sequence of smaller, simpler transformations, which is thought to be the key to producing high-quality samples. Each of these transformations is typically carried out by a separate step in the sampling process, and as a result sampling takes many steps to iterate over all the transformations. Iterative sampling is, however, computationally expensive, since it involves multiple forward passes through the model. If we can achieve high-quality generation without the cost of iterative sampling, we can make image generation both cheaper and simpler. 

To this end, we start with a single-step model and add only a minimal set of modifications that are necessary to achieve competitive performance. To devise the modifications, we find the mechanisms behind SI models that are truly important, abstract them at a high level to enable application to the single-step setting and add them to our single-step model. Rather than viewing SI models as performing iterative sampling, we take a testing-centric view. We hypothesize that the ingredient that powers their success is per-stage supervision and add it to our single-step model. Our second contribution is specific to the matching phase in IMLE, which can occasionally produce poor matches due to the randomness in the sampling process. We introduce a simple change to the algorithm that makes it robust to these outliers. In line with our minimalist design philosophy, we use a ConvNeXt-based architecture~\cite{liu2022convnet2020s} rather than the transformer-based architectures that now dominate. 

Together, these design choices yield \textbf{Ro}bust \textbf{M}ulti-\textbf{S}tage IMLE (\methodshort{}), a single-step generative model that combines per-stage supervision with a robust training loss. 
Across three diverse datasets, our model attains competitive FID and strong precision and recall. In pixel space, \methodshort{} attains the highest precision and recall compared to baselines on CIFAR-10 and CelebA-HQ. For latent generation on ImageNet 256, it reaches an FID of 2.56 in a single forward pass, with 54\% fewer parameters and 250$\times$ fewer function evaluations than DiT-XL/2 and SiT-XL/2. Through our results, we demonstrate that a simple single-step generative model can indeed retain the sample quality and diversity commonly exhibited by popular SI models without resorting to decomposing generation into gradual transformations.

\begin{figure*}
  \begin{center}
      \includegraphics[width=0.9\linewidth]{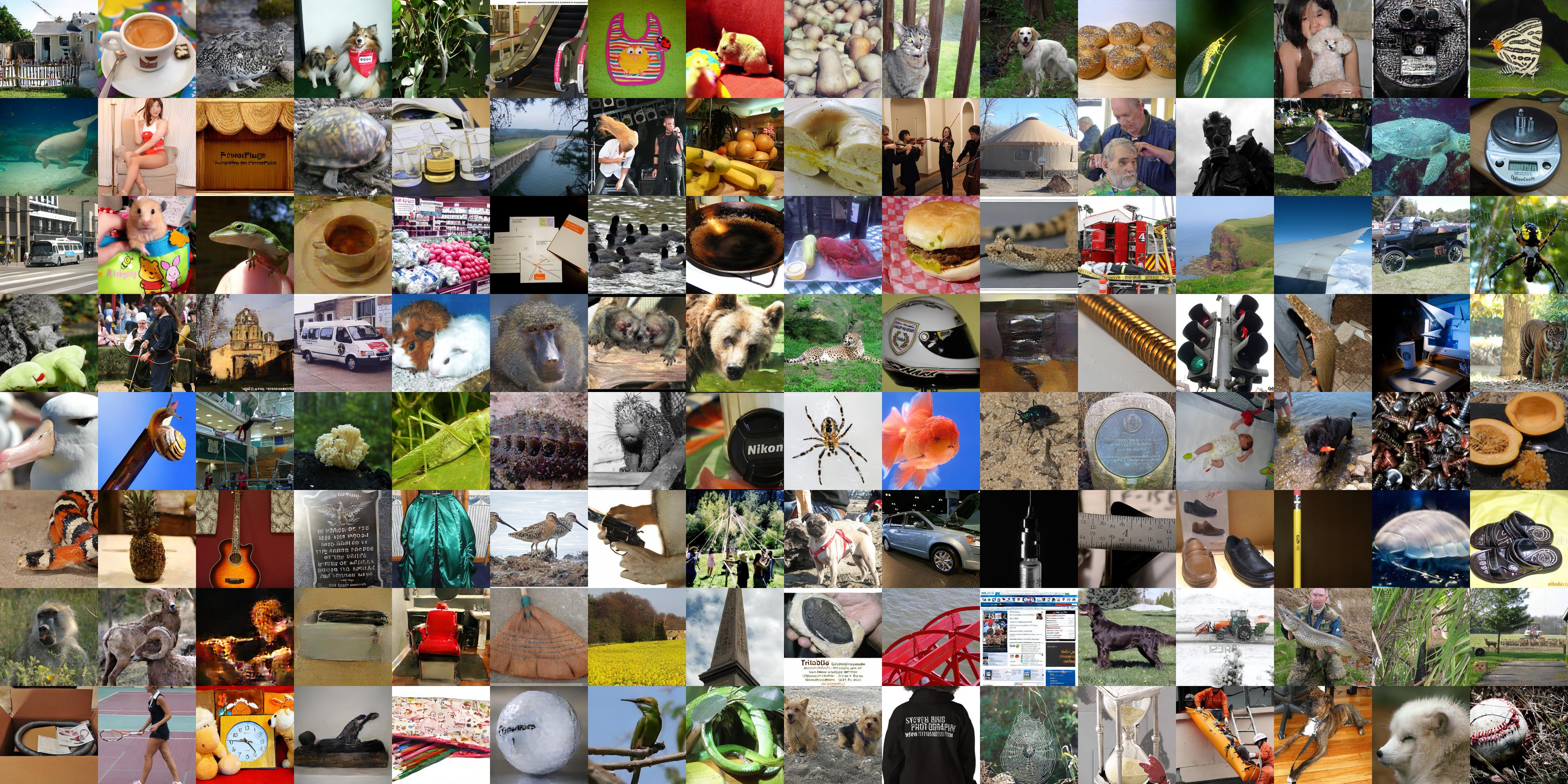}
  \end{center}
  \caption{Random samples from our class-conditional model trained on ImageNet}

  \label{fig:imagenet-samples}
\end{figure*}

\section{Related Works} 
\label{related-works}

\textbf{One-step generative models}
GANs remain strong one-step image generators~\cite{brock2018large, stylegan-xl, styleswin, styleformer} but are difficult to train \cite{wgans,training-gans-1} and prone to mode collapse \cite{mode-collapse-1, mode-collapse-2}. 
A separate line of research tries to make iterative stochastic interpolant models fast by reducing the number of sampling steps, through distillation~\cite{salimans2022progressivedistillationfastsampling}, consistency training~\cite{song2023consistencymodels}, or redesigned objectives such as shortcut models~\cite{frans2024stepdiffusionshortcutmodels}, MeanFlow~\cite{meanflow, geng2025improved}, and inductive moment matching~\cite{zhou2025inductive}. In recent years, IMLE~\cite{li2018implicitmaximumlikelihoodestimation} has received increasing attention for its mode-covering objective and strong performance in few-shot image synthesis~\cite{vashist2024rejectionsamplingimledesigning, adaptiveIMLE}. IMLE has also been adopted beyond image generation, in areas such as visuomotor policy learning~\cite{rana2025imlepolicyfastsample}, trajectory planning~\cite{imle-trajectory}, and shape completion~\cite{imle-shape-completion}.

\section{Background}

\subsection{IMLE}

The goal of generative modeling is to transform a simple, easy-to-sample prior distribution into the complex data distribution we wish to model. One way to approach this problem is to learn the transformation directly, using a neural generator that maps samples from the prior to data samples in a single pass. In general, this approach is difficult because the likelihood of the data under the generator is intractable (refer to Appendix \ref{appendix-mle} for more details). Implicit Maximum Likelihood Estimation (IMLE) ~\cite{li2018implicitmaximumlikelihoodestimation,adaptiveIMLE} provides a workaround by implicitly maximizing the likelihood without requiring knowledge of its explicit form. 

Formally, IMLE learns a direct mapping $f_\theta$ from a prior $\pi(\mathbf{z})$ to the data distribution $p_\mathrm{data}(\mathbf{x})$ by minimizing the following objective:
\begin{equation}
    \theta_{\text{IMLE}} = \argmin_\theta \mathbb{E}_{\mathbf{z}_1,\ldots,\mathbf{z}_m \sim \pi(\mathbf{z})}
    \left[
        \sum_{i=1}^n \min_{j \in [m]}  \; d\bigg( \mathbf{x}_i, f_\theta(\mathbf{z}_j) \bigg)
    \right],
\label{eqn:method-imle}
\end{equation}
where $d(\cdot,\cdot)$ is a distance function.
To optimize the objective, the IMLE algorithm alternates between two phases. In the matching phase, $m$ latent codes $\mathbf{z}_1, \ldots, \mathbf{z}_m$ are sampled from the prior and each real image $\mathbf{x}_i$ is matched to its nearest generated sample in the set $\{f_\theta(\mathbf{z}_1), \ldots, f_\theta(\mathbf{z}_m)\}$. This establishes a coupling between $\pi(\mathbf{z})$ and $p_\mathrm{data}(\mathbf{x})$. Next, in the optimization phase, the parameters $\theta$ are updated to bring each generated sample closer to its matched real image. We now take a look at SI models to understand how their construction differs from IMLE, and whether any of their properties can be leveraged in a single-step model.

\subsection{Flow Matching and Testing-Centric View}
\label{sec:testing-centric}

Stochastic interpolant (SI) methods, such as diffusion and flow matching, are commonly understood to perform iterative denoising. At training time, SI methods train a model, commonly known as a denoising network, to convert a noisy image to a clean image. At test time, the denoising network is repeatedly applied to convert pure noise to a generated sample. 
For the purpose of this section, we consider flow matching as an example of SI methods. Given the data distribution $p_\mathrm{data}$ and prior $p_{T}$ distribution, the general objective takes the following form:

\begin{equation}
    \mathcal{L}(\theta) = \mathbb{E}_{t \sim \mathcal{U}(0, T],\, \mathbf{x}_0 \sim p_\mathrm{data},\, \mathbf{x}_T \sim p_T,\, \mathbf{x}_t \sim q_t(\mathbf{x}_t \vert \mathbf{x}_0, \mathbf{x}_T)} \left[ \ell_t(g^t(\mathbf{x}_0, \mathbf{x}_t), g^t_{\theta}(\mathbf{x}_t) ) \right]
\label{eqn:ind-loss}
\end{equation}

The function $g^t_{\theta}$ is the model's prediction at time $t$, $g^t$ is the supervision target it is trained to match, and $\ell_t$ compares the two. $q_t$ characterizes the interpolation between $p_\mathrm{data}$ and $p_T$. 

A key distinguishing feature of SI methods is the iterative application of the denoising network, and this is what SI methods are believed to owe their success to. 
This need for multiple steps creates a mismatch between training and test; the denoising network is applied once for each sample during training, whereas it is applied iteratively to generate a sample during test. We consider an alternative view of this mismatch. Rather than viewing the model as one instance of the denoising network, we unroll the different iterations of the sampling procedure and view the model as a composition of the different iterations. Under this view, at test time, the model is just applied once to convert pure noise to a generated sample, and at training time, only a part of the model is evaluated at a time. This is in some sense the flip side of the conventional view; under the conventional view, the test time procedure is what deviates from the norms, whereas under the alternative view, the training time procedure is what deviates from the norms. Hence, we will term the conventional view the \emph{training-centric view} and the alternative view the \emph{testing-centric view}. 

We now concretely construct the model under the testing-centric view. Consider a deterministic sampler in an SI method, such as those used in DDIM or flow matching. Starting from Gaussian noise, the sampler applies a succession of updates parameterized by the denoising network evaluated at different time steps. More precisely, the sampler starts by sampling $\mathbf{x}_T \sim p_T(\mathbf{x})$ and produces a sequence $\mathbf{x}_T \to \mathbf{\hat{x}}_{T-1} \to \cdots \to \mathbf{\hat{x}}_0$, where at each iteration $\mathbf{\hat{x}}_{t-1}$ is obtained from $g^t_{\theta}(\mathbf{\hat{x}}_t)$. 

Under the testing-centric view, the model is the composition of all the updates done by the sampler, namely $g^T_{\theta}, \ldots, g^1_{\theta}$. %
\begin{equation}
    \mathbf{\hat{x}}_0 \;=\; \underbrace{g^1_{\theta} \circ g^2_{\theta} \circ \cdots \circ g^T_{\theta}}_{h_{\theta}}(\mathbf{x}_T),
    \label{eqn:diffusion-composition}
\end{equation}
This allows us to view the reverse process as a \emph{single} compositional model $h_{\theta}$, that maps a sample from the prior $\mathbf{x}_T$ to a data sample $\mathbf{\hat{x}}_0$. Hence, what the training-centric view treats as an iteration over time steps is now viewed as an iteration over depth in the compositional model.
Under this view, each $\mathbf{\hat{x}}_{t-1} = g^t_{\theta}(\mathbf{\hat{x}}_t)$ is an intermediate layer of $h_{\theta}$. Let us revisit the training objective. If we restrict the support of $t$ to integer time steps in Equation~\ref{eqn:ind-loss}, we can rewrite the objective as:
\begin{equation}
    \tilde{\mathcal{L}}(\theta) = \frac{1}{T}\sum_{t=1}^{T}\mathbb{E}_{\mathbf{x}_0 \sim p_\mathrm{data},\, \mathbf{x}_T \sim p_T,\, \mathbf{x}_t \sim q_t(\mathbf{x}_t \vert \mathbf{x}_0, \mathbf{x}_T)} \left[ \ell_t(g^t( \mathbf{x}_0, \mathbf{x}_t), g^t_{\theta}(\mathbf{x}_t) ) \right]
\label{eqn:ind-loss-2}
\end{equation}

\begin{figure*}[t]
  \centering
  \includegraphics[width=\textwidth]{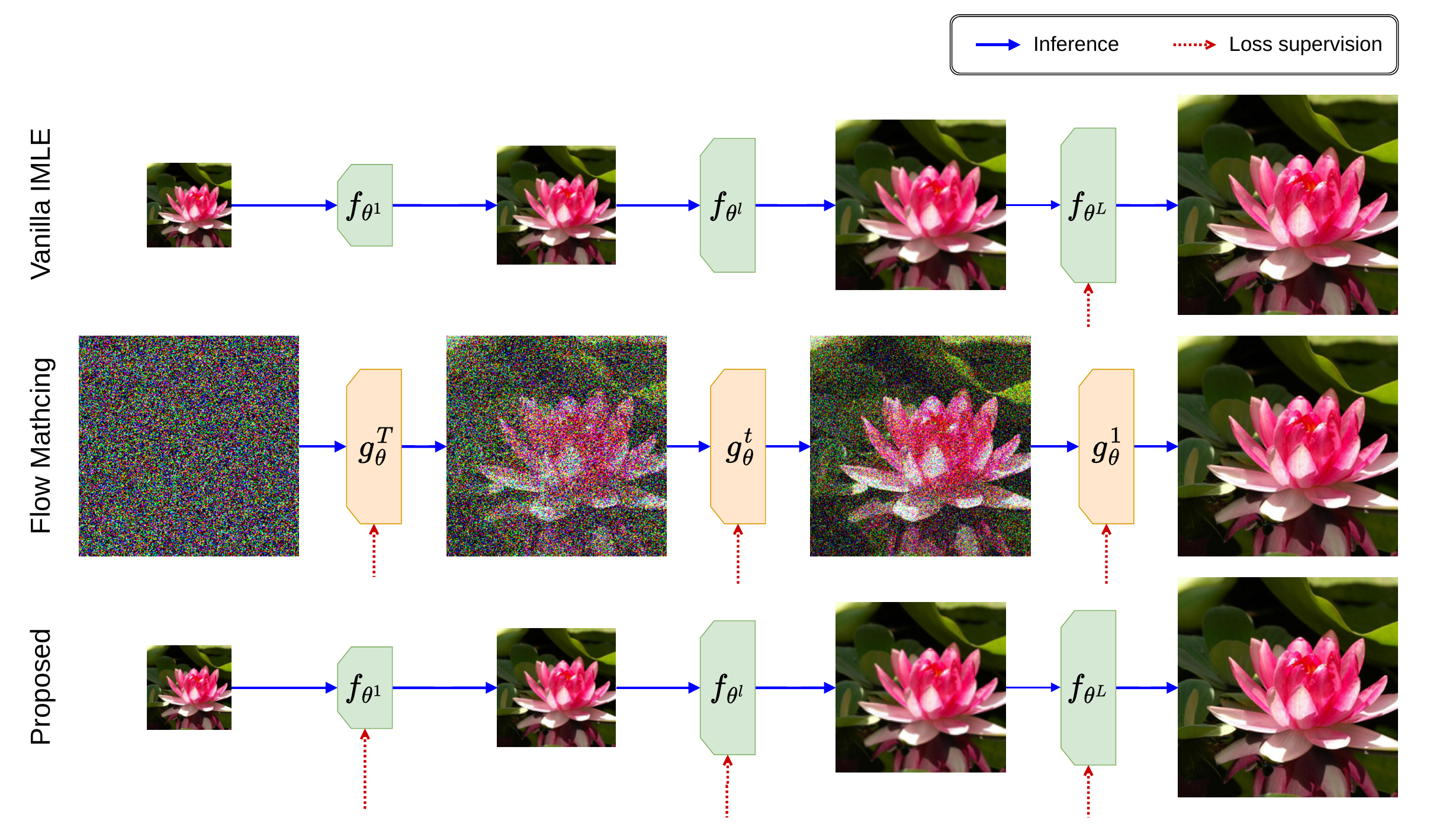}
  \caption{\textbf{Multi-stage supervision:} Each row is a generator drawn as a composition of stages; \textcolor{blue}{blue} arrows are the inference path and \textcolor{red}{red} dotted arrows show where the loss is applied. Vanilla IMLE (top) supervises only its final stage, whereas flow matching (middle) supervises every denoising stage. Our proposed method (bottom) retains IMLE's architecture and one-step inference while extending direct supervision to every stage.}
  \label{fig:composition}
\end{figure*}

We emphasize how the compositional model $h_{\theta}$ is trained by the loss in Equation~\ref{eqn:ind-loss-2}, noting that each stage $g^t_{\theta}$ in the composition is directly supervised. Under the testing-centric view, the training objective therefore performs \emph{intermediate per-stage supervision}: every intermediate layer $g^t_{\theta}$ of the compositional model $h_{\theta}$ receives a direct training signal (Figure~\ref{fig:composition}). This training-time property, rather than iterative sampling, is what we believe is the reason for high-quality generation in SI models.

\section{Method}

% Unlike SI methods, IMLE chooses the pairing rather than leaving it arbitrary: its matching phase pairs each data point with its nearest generated sample. As a consequence, the supervision target is a single image rather than an average, which means a single step suffices. We therefore take IMLE as our starting point and improve how it is trained: we add the supervision it lacks, and we address a cost that its matching phase introduces.

Our goal in this paper is to improve IMLE, a single-step generator, so that it can match the sample quality of modern generative models that generate over many sequential steps. The two contributions presented in the paper improve on the optimization and matching phases of IMLE, respectively. The first contribution, \emph{multi-stage supervision}, improves the optimization phase of IMLE by providing a direct training signal to each stage of the generator. Our second contribution, \emph{robust training loss}, improves the matching phase of IMLE by making it more robust to outliers. We describe them in greater detail below.

\subsection{Optimization Phase Improvement: Adapting Multi-Stage Supervision for IMLE}

In the previous section, we presented the SI sampler as a single compositional model $h_\theta$. We can now contrast this compositional model with the single-step generation setup of vanilla IMLE. The SI objective supervises every stage of the composition, whereas vanilla IMLE supervises only the final output of $f_\theta$. Supervision at every stage is the property we want to carry over. The unrolled view introduced in Section \ref{sec:testing-centric} applies naturally to IMLE: its generator can also be seen as a prior-to-data network composed of stages, only executed in a single forward pass rather than through iterative sampling. What it lacks is \emph{multi-stage supervision}, which we add through a change to the training objective.

\textbf{Compositional scaffold} 
We choose to build the IMLE generator $f_\theta$ as a stack of upsampling blocks. We adopt this design deliberately; similar to Section~\ref{sec:testing-centric}, it lets us write $f_\theta$ as a composition
$$f_\theta = f_{\theta^{L}} \circ f_{\theta^{L-1}} \circ \cdots \circ f_{\theta^{1}},$$
in which each stage $f_{\theta^{l}}$ operates at a progressively higher resolution, with per-stage parameters $\theta = (\theta^{1}, \ldots, \theta^{L})$. 
This recasts IMLE in a form similar to the diffusion construction, with each $f_{\theta^{l}}$ now analogous to $g^t_{\theta}$, only indexed by resolution rather than noise level. 

The architecture of $f_\theta$ in existing IMLE methods \cite{vashist2024rejectionsamplingimledesigning,adaptiveIMLE} is compositional, but the training is not. The loss in Equation~\ref{eqn:method-imle} is computed only at the final output $f_\theta(\mathbf{z})$, which directly supervises only stage $L$. Stages $1, \ldots, L-1$ receive a training signal only through gradients that flow back from this final loss. We address this by supervising every stage directly, at the resolution at which it operates.

Concretely, two things are needed: a way to read out a prediction at each stage, and a way to obtain a target at the matching resolution. We add a per-stage output head (see Figure~\ref{fig:composition}), so that stage $f_{\theta^{l}}$ outputs an image $\tilde{\mathbf{x}}^{l}$ at its own resolution. We obtain the target by downsampling the real image $\mathbf{x}_i$ to the same resolution with a kernel $K_l$. At each stage $l$, the per-stage loss term compares $\tilde{\mathbf{x}}^{l}$ against $K_l(\mathbf{x}_i)$. Each per-stage loss is averaged over the elements at that stage's resolution, so the $L$ terms are on a comparable scale. The training objective is the average of all $L$ per-stage loss terms.

\begin{algorithm}
\caption{\methodshort{} training procedure}
\label{alg:our_proc}
\begin{algorithmic}[1]
\Require Dataset $\{\mathbf{x}_i\}_{i=1}^n$, generator $f_\theta = f_{\theta^{L}} \circ \cdots \circ f_{\theta^{1}}$ with parameters $\theta = (\theta^{1}, \ldots, \theta^{L})$, downsampling kernel $K_l(\cdot)$, distance function $d(\cdot, \cdot)$ averaged over the elements of its arguments
\State Initialize $\theta$ to a random vector
\For{$t = 1$ \textbf{to} $T$}
    \State Draw i.i.d.\ latent codes $\mathbf{z}_1, \ldots, \mathbf{z}_m \sim \mathcal{N}(0, I)$
    \State $\tilde{\mathbf{x}}_j \gets f_\theta(\mathbf{z}_j)$ for all $j \in [m]$ \Comment{Generate samples}
    \State $\sigma(i) \gets \arg\min_{j \in [m]} d\!\left(\mathbf{x}_i,\; \tilde{\mathbf{x}}_j\right) \quad \forall i \in [n]$ \Comment{Nearest-neighbour matching}
   \For{each minibatch $S \subseteq [n]$}
        \State$\tilde{\mathbf{x}}_{\sigma(i)}^{l} \gets (f_{\theta^{l}} \circ \cdots \circ f_{\theta^{1}})(\mathbf{z}_{\sigma(i)})$ for all $i \in S$, $l \in [L]$ \Comment{Per-stage outputs}
        \State $\mathcal{J} \gets 0$
        \For{$l = 1$ \textbf{to} $L$}
            \State $\mathcal{J} \gets \mathcal{J} + \frac{1}{|S|} \sum_{i \in S} d\!\left(K_l(\mathbf{x}_i),\; \tilde{\mathbf{x}}_{\sigma(i)}^{l}\right)$ \Comment{Per-stage supervision}
        \EndFor
        \State $\theta \gets \theta - \frac{\eta}{L}\, \nabla_\theta \mathcal{J}$ \Comment{Gradient step}
    \EndFor
\EndFor
\State \Return $\theta$
\end{algorithmic}
\end{algorithm}

\begin{figure*}[]
  \centering
  \begin{subfigure}[t]{0.32\textwidth}
    \includegraphics[width=\linewidth]{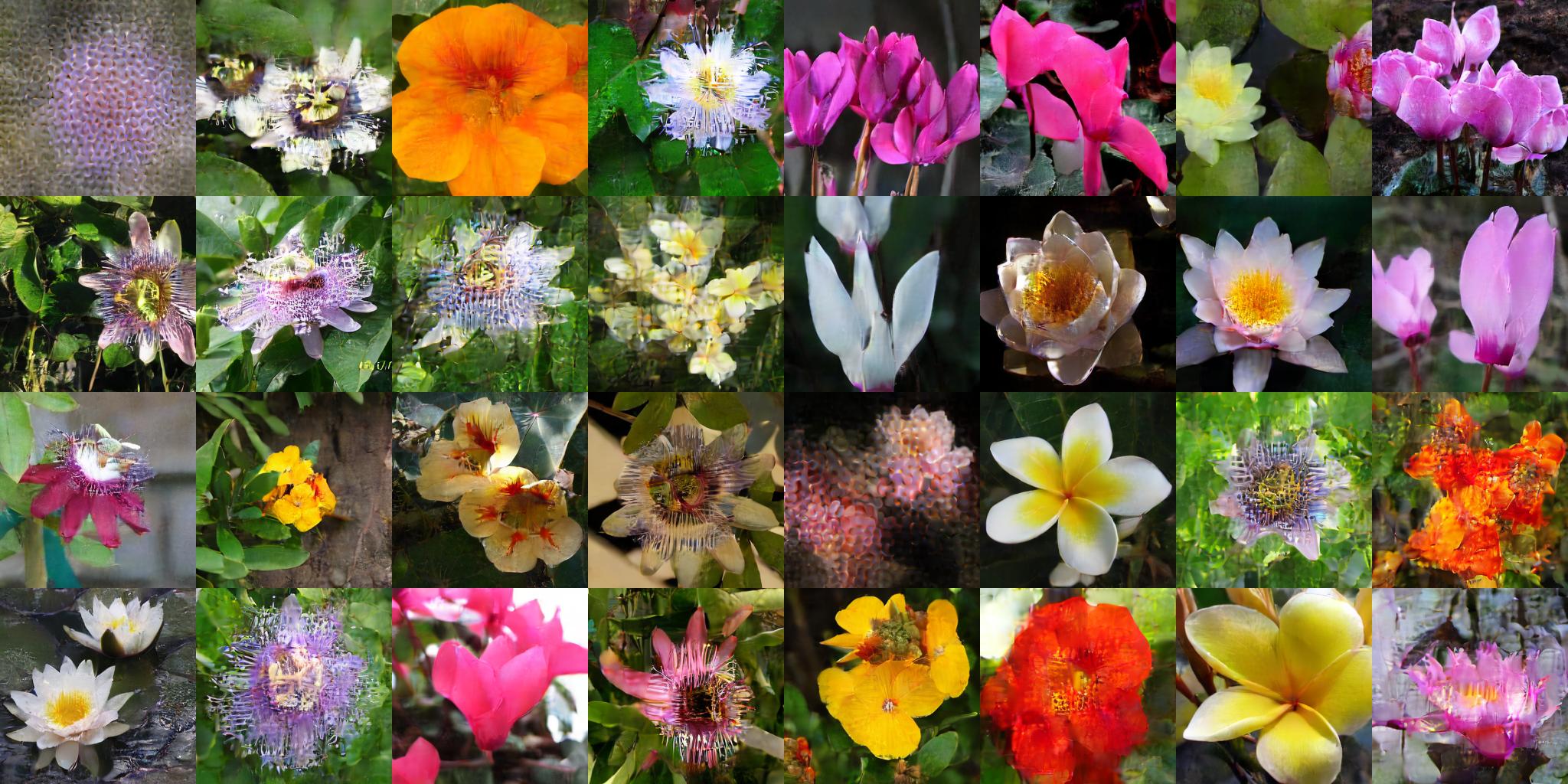}
    \caption{\textbf{Conf-A}: Vanilla IMLE}
  \end{subfigure}
  \hfill
  \begin{subfigure}[t]{0.32\textwidth}
    \includegraphics[width=\linewidth]{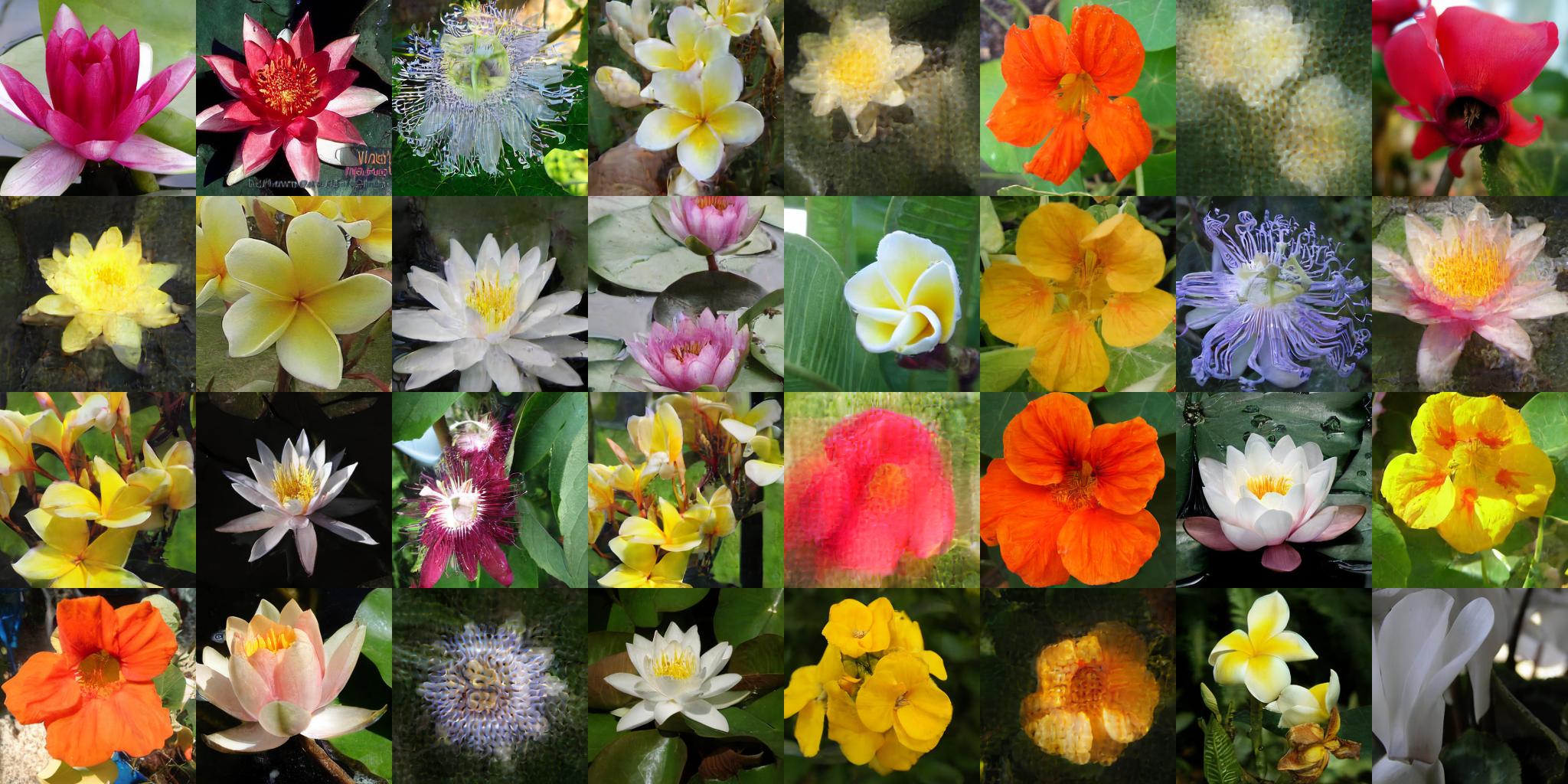}
    \caption{\textbf{Conf-B}: multi-stage supervision}
  \end{subfigure}
  \hfill
  \begin{subfigure}[t]{0.32\textwidth}
    \includegraphics[width=\linewidth]{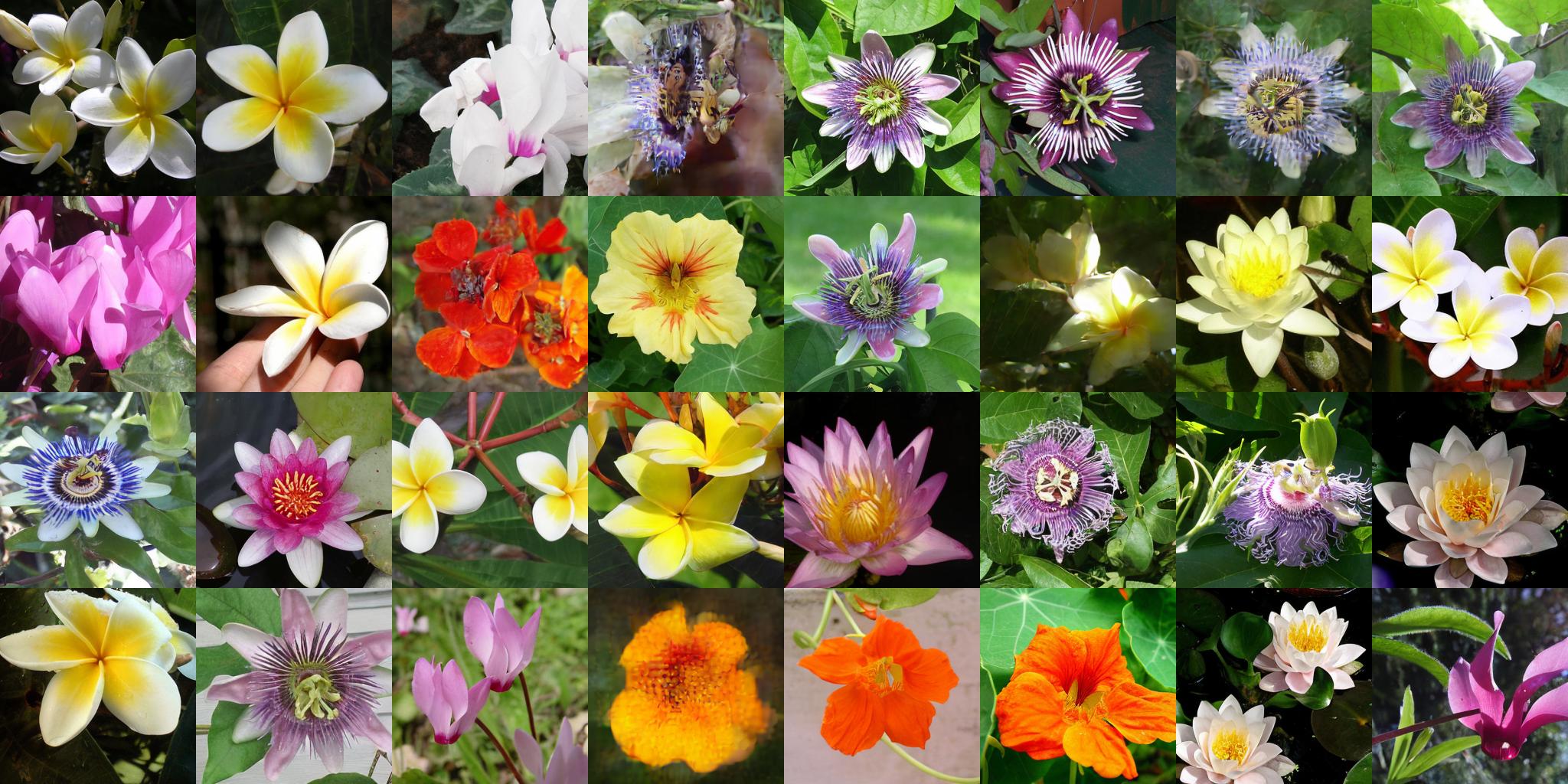}
    \caption{\textbf{Conf-E}: multi-stage supervision + Geman-McClure loss}
  \end{subfigure}
  \caption{\textbf{Qualitative ablation}: Random samples from IMLE models trained on Oxford Flowers (256$\times$256) for selected configurations in Table~\ref{tab:method-ablation}. The multi-stage supervision in Conf-B substantially improves sample quality over the vanilla IMLE in Conf-A. The robust loss in Conf-E further improves sample quality, yielding sharper and more realistic images.}
  \label{fig:ablation-samples}
\end{figure*}

We test this new training procedure on Oxford Flowers~\cite{oxford-flowers}, an image dataset of roughly 1000 images at $256 \times 256$ resolution. We use this dataset for the ablation because it is small enough to train and iterate over quickly, yet visually diverse enough that mode collapse is easy to spot in the generated samples. Table~\ref{tab:method-ablation} reports the results and Figure~\ref{fig:ablation-samples} shows random samples from each configuration: our proposed training paradigm with multi-stage supervision (B) substantially outperforms vanilla IMLE (A). Additional experiment details can be found in Appendix~\ref{appendix-ablation-flowers}. 

\begin{table}[h]
\centering
\caption{Ablation study on Oxford Flowers \cite{oxford-flowers} (256$\times$256), evaluated at the same training iteration. We report the mean $\pm$ standard deviation over 3 seeds. Row B adds multi-stage supervision to the baseline. Rows C-E each replace the loss function in Row B with a different robust loss.}
\label{tab:method-ablation}
\small
\begin{tabular}{clccc}
\toprule
& \textbf{Configuration} & FID $\downarrow$ & Precision $\uparrow$ & Recall $\uparrow$ \\
\midrule
A & Vanilla IMLE                                      & 62.61 $\pm$ 2.28 & 0.45 $\pm$ 0.03 & 0.46 $\pm$ 0.01 \\
B & \ \ + multi-stage supervision                 & 31.24 $\pm$ 1.08 & 0.79 $\pm$ 0.01 & 0.72 $\pm$ 0.02 \\
\midrule
C & (B) + Cauchy loss                                  & 19.80 $\pm$ 0.58 & 0.88 $\pm$ 0.01 & \textbf{0.74 $\pm$ 0.01} \\
D & (B) + Pseudo-Huber loss                            & 16.49 $\pm$ 0.09 & 0.91 $\pm$ 0.00 & 0.72 $\pm$ 0.02 \\
E & (B) + Geman-McClure loss \textbf{(ours)}           & \textbf{13.90 $\pm$ 0.08} & \textbf{0.93 $\pm$ 0.00} & 0.73 $\pm$ 0.02 \\
\bottomrule
\end{tabular}
\end{table}

\subsection{Matching Phase Improvement: Robust Loss for Mismatched Pairs}

\begin{figure*}
  \centering
  \small
  \begin{tabular}{@{}m{0.10\linewidth}@{\hspace{4pt}}m{0.88\linewidth}@{}}
    \centering\small Ground\newline Truth\newline Image &
    \begin{tikzpicture}
      \node[inner sep=0] (img) {\includegraphics[width=\linewidth]{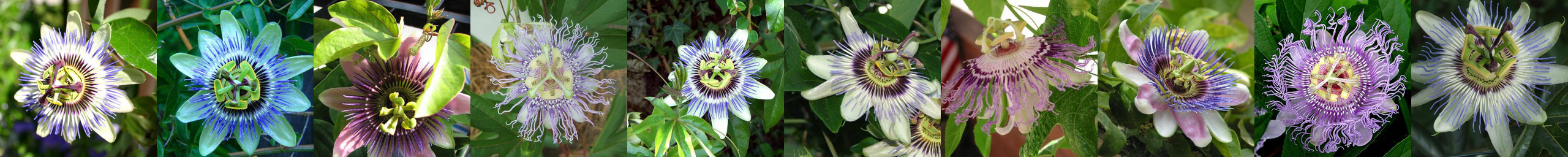}};
      \draw[orange, line width=1.5pt]
        ($(img.south west)!0.6!(img.south east)$) rectangle ($(img.north west)!0.7!(img.north east)$);
    \end{tikzpicture} \\[4pt]
    \centering\small Current \newline Nearest\newline Neighbour &
    \begin{tikzpicture}
      \node[inner sep=0] (img) {\includegraphics[width=\linewidth]{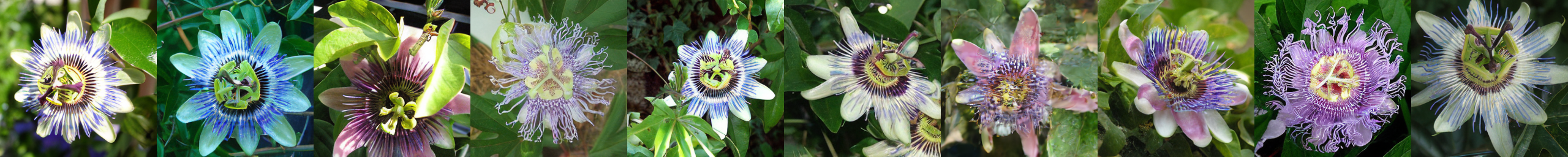}};
      \draw[orange, line width=1.5pt]
        ($(img.south west)!0.6!(img.south east)$) rectangle ($(img.north west)!0.7!(img.north east)$);
    \end{tikzpicture} \\[4pt]
    \centering\small Previous \newline Nearest\newline Neighbour &
    \begin{tikzpicture}
      \node[inner sep=0] (img) {\includegraphics[width=\linewidth]{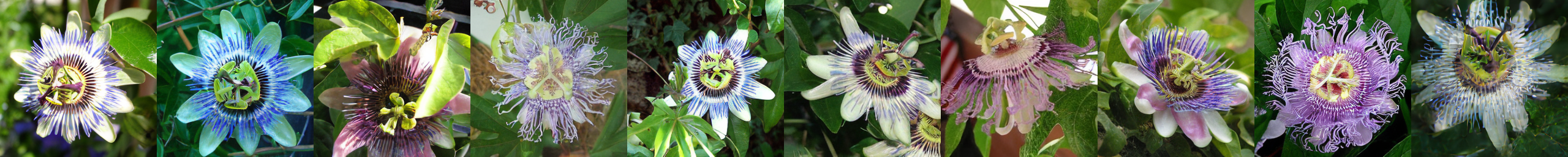}};
      \draw[orange, line width=1.5pt]
        ($(img.south west)!0.6!(img.south east)$) rectangle ($(img.north west)!0.7!(img.north east)$);
    \end{tikzpicture} \\
  \end{tabular}
  \caption{\textbf{Mismatch in nearest-neighbour pairing.} Ground-truth image (top) and its nearest neighbours among generated samples in the current round (middle) and the previous round (bottom). For most images, both are close to the ground truth. For the \textcolor{orange}{highlighted} image, the current-round nearest neighbour does not resemble the ground truth.}
  \label{fig:mismatch}
\end{figure*}
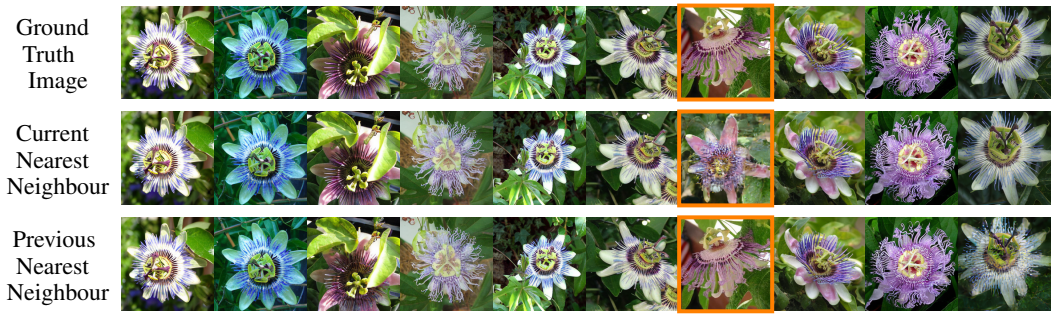

SI models interpolate between the noise distribution and the empirical data distribution along a predetermined trajectory, and so the flow between them does not need to be recomputed during training. On the other hand, in a single-step model like IMLE, the model distribution changes during training, so the matching between model samples and ground truth data points needs to be recomputed. Due to randomness in the sampling, there may be no samples drawn from a mode of the model distribution, and as a result, a ground truth data point that could otherwise be explained by that mode would not be well matched. Note that this is different from mode collapse: the poor matching results from randomness in the sampling, not a deficiency in the model distribution. 

Figure~\ref{fig:mismatch} shows this phenomenon, where the ground truth image (top row) is shown alongside its nearest-neighbour matches in the current round (middle) and the previous round (bottom). For most images, both matches found in consecutive rounds of sampling are close to the ground truth, indicating that the generator has learned these targets well. For the highlighted image, however, the current-round match does not look like the ground truth, while the previous-round match does. %

The standard squared $\ell_2$ loss lets these mismatches dominate training, since its gradient with respect to the generated sample is proportional to the residual. A mismatched pair has a large residual and therefore contributes a correspondingly large gradient. The problem is, in essence, one of outliers: a small number of large-residual pairs distort the gradient computed from a much larger pool of well-matched pairs. 

The classical remedy in robust statistics is to replace the squared loss with a function that grows sub-quadratically in the residual. We adopt this approach: we replace $d(\cdot, \cdot)$ with a robust loss function. When a mismatch occurs and produces a large residual, the robust loss reduces its influence on the gradient, allowing the remaining well-matched pairs to dominate the update. We apply the robust loss at each decoder stage and ablate three candidates~\cite{barron-robust}: Cauchy, Pseudo-Huber, and Geman-McClure. As shown in Table~\ref{tab:method-ablation}, all three improve performance over the non-robust loss, with Geman-McClure performing best. Figure~\ref{fig:ablation-samples} shows that Geman-McClure with multi-stage supervision also yields the best qualitative performance, producing sharper and more realistic images than the other configurations.

\subsection{Round-Trip Rejection for Latent-Space Generation}

For high-resolution datasets, a common paradigm is to generate in the latent space \cite{vahdat2021scorebasedgenerativemodelinglatent, rombach2022highresolutionimagesynthesislatent} of a frozen, pretrained autoencoder and decode the generated latents into images. In the literature on SI models and GANs, inference-time tradeoffs of quality against diversity are popular and are enabled by techniques such as classifier-free guidance (CFG) \cite{cfg} and the truncation trick \cite{brock2018large}. To enable the same tradeoff, we design a simple technique, which we call round-trip rejection. 

Latent space generation introduces a distribution mismatch: the autoencoder faithfully reconstructs only the images it was trained on, but our generator is not constrained to stay within that distribution and occasionally produces images the autoencoder does not represent well, which decode into visibly degraded samples. We add an inference-time step that identifies and discards these samples. We detect them using a \emph{round-trip cost}: the distance between a generated image and the image obtained by encoding and then decoding it again. This round trip projects an image onto the distribution the autoencoder can represent, so a well-represented image changes little (low cost) while a poorly-represented one changes substantially (high cost). We reject generated samples whose round-trip cost is high. %
We include further implementation details in Appendix~\ref{appendix-details}.

\textbf{Putting it together.} Combining multi-stage supervision with the robust Geman-McClure loss yields our final training recipe, which we refer to as \methodname{} (\methodshort{}). The architecture and inference path are unchanged from vanilla IMLE; the changes are entirely in the training objective. Algorithm~\ref{alg:our_proc} states the full procedure.

\section{Experiments}
\label{experiments}

\begin{figure*}
  \begin{center}
    \includegraphics[width=0.9\linewidth]{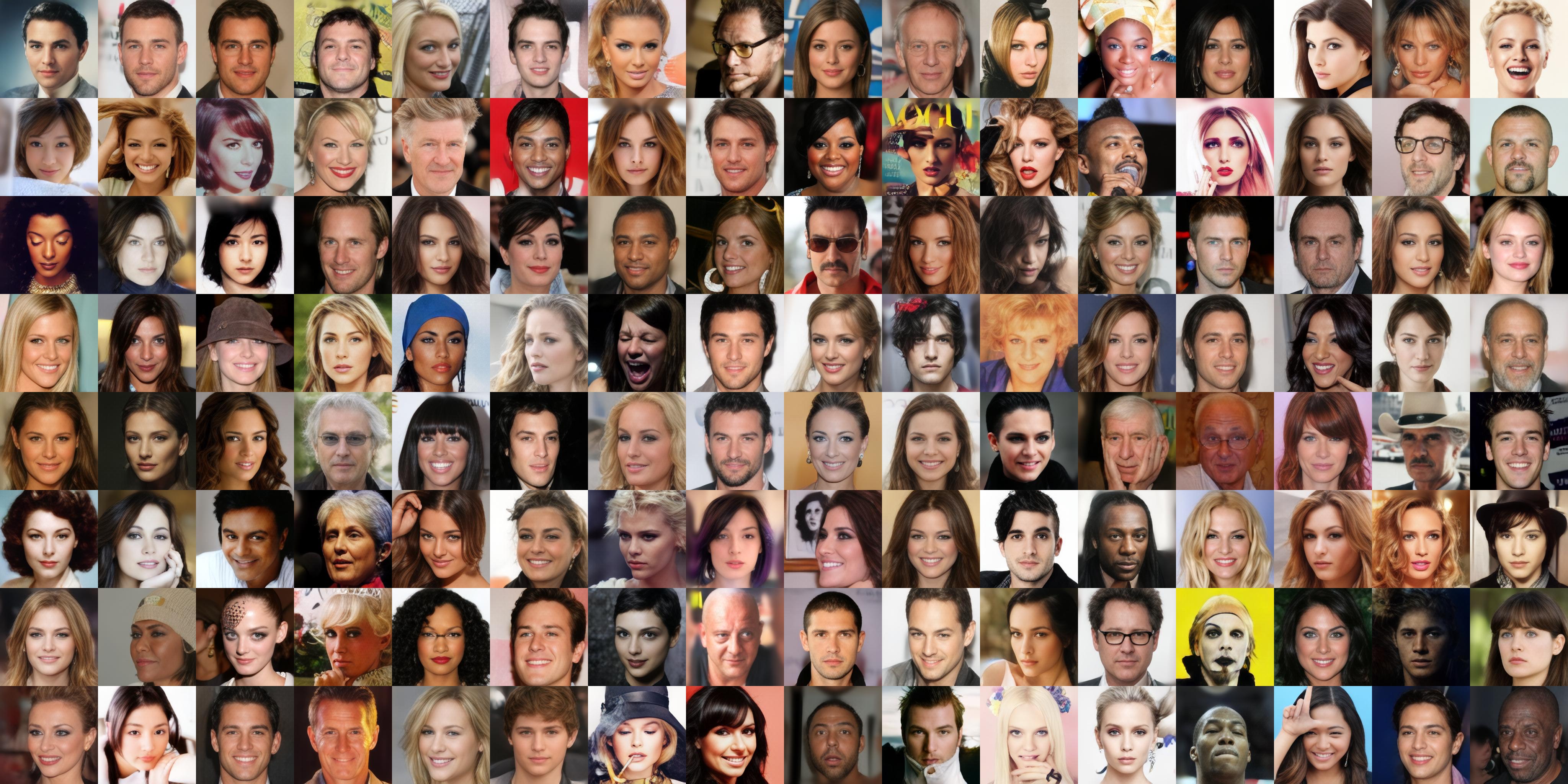}
  \end{center}

  \caption{Random samples from our unconditional model trained on CelebA-HQ}

  \label{fig:celeba-samples}
\end{figure*}

\textbf{Datasets} We demonstrate the performance of our method on three datasets: unconditional generation on CIFAR-10 (50K images, $32\times32$ resolution)~\cite{krizhevsky2009learning} and CelebA-HQ (30K images, $256\times256$ resolution)~\cite{karras2018progressivegrowinggansimproved}, and class-conditional generation on ImageNet 256~\cite{imagenet}.

\textbf{Evaluation metrics} 
We use the FID-50k metric~\cite{fid} to assess the perceptual quality of the generated images, as is standard in prior work. Since FID is not always a perfect indicator of perceptual quality~\cite{rethinking-fid}, we additionally evaluate fidelity and coverage by computing precision and recall using the metrics defined by Kynk\"a\"anniemi et al.~\cite{kynkaanniemi2019improved}. More details on the evaluation metrics are provided in Appendix~\ref{appendix-eval}.

\textbf{Implementation details} Our network architecture comprises a fully-connected mapping network inspired by~\cite{karras2019stylebasedgeneratorarchitecturegenerative} and a synthesis network constructed using ConvNeXt~\cite{liu2022convnet2020s} modules. We use bicubic interpolation for downsampling images/latents.
We use the AdamW optimizer~\cite{adamw} with $\beta_{1},\beta_{2}$ set to 0.9. Our method works in both pixel and latent space; to demonstrate that, we train our model in pixel space for CIFAR-10 and CelebA-HQ, and in the EQ-VAE~\cite{eqvae} latent space for ImageNet 256.

\begin{table*}[t]
  \centering
  \begin{minipage}[t]{0.50\textwidth}
    \centering
    \caption{Results for unconditional generation on CIFAR-10}
    \label{tab:cifar10-results}
    \small
    \setlength{\tabcolsep}{3pt}
    \begin{tabular}{lrSSSS}
      \toprule
      \textbf{Method} & \textbf{NFE} & \textbf{FID $\downarrow$} & \textbf{IS $\uparrow$} & \textbf{Pr. $\uparrow$} & \textbf{Re. $\uparrow$} \\
      \midrule
      DDIM~\citep{song2022denoisingdiffusionimplicitmodels} & 100 & 11.18              & 8.62                & 0.619               & 0.567               \\
      CT~\cite{song2023consistencymodels}                   & 1   & 8.79               & 8.61                & {\underline{0.70}} & 0.419               \\
      Mean Flows~\cite{meanflow}                            & 1   & {\textbf{2.87}} & \underline{10.11}               & 0.687               & 0.586               \\
      IMM~\cite{zhou2025inductive}                          & 1   & 3.16               & 10.10               & 0.659               & \underline{0.60}               \\
      StyleFormer~\cite{styleformer}                        & 1   & \underline{2.88}               & {\textbf{10.26}}    & 0.643               & 0.540               \\
      \midrule
      \textit{\methodshort} (Pixel)                                    & 1   & 3.93               & {{10.01}} & {\textbf{0.91}}     & {\textbf{0.80}}     \\
      \bottomrule
    \end{tabular}
  \end{minipage}\hfill
  \begin{minipage}[t]{0.45\textwidth}
    \centering
    \caption{Results for unconditional generation on CelebA-HQ}
    \label{tab:celeba}
    \small
    \setlength{\tabcolsep}{3pt}
    \begin{tabular}{@{}lcccc@{}}
      \toprule
      \textbf{Method} & \textbf{NFE} & \textbf{FID} $\downarrow$ & \textbf{Pr.} $\uparrow$ & \textbf{Re.} $\uparrow$ \\
      \midrule
      Shortcut~\cite{frans2024stepdiffusionshortcutmodels} & 1   & 23.65            & \underline{0.76} & 0.16             \\
      Shortcut~\cite{frans2024stepdiffusionshortcutmodels} & 128 & 10.79            & 0.71             & \underline{0.41} \\
      StyleSwin~\cite{styleswin}                           & 1   & \textbf{5.32}    & 0.70             & 0.39             \\
      \midrule
      \textit{\methodshort} (Pixel)                                   & 1   & \underline{6.70} & \textbf{0.96}    & \textbf{0.61}    \\
      \bottomrule
    \end{tabular}
  \end{minipage}
\end{table*}

\begin{table}[ht]
\centering
\caption{Results for class-conditional image generation on ImageNet 256}
\label{tab:imagenet-results}
\resizebox{\textwidth}{!}{%
\begin{tabular}{lrrSSS}
\toprule
\textbf{Model}
  & {\textbf{Params (M)}}
  & {\textbf{NFE} }
  & {\textbf{FID} $\downarrow$}
  & {\textbf{Pr.} $\uparrow$}
  & {\textbf{Re.} $\uparrow$} \\
\midrule
BigGAN\cite{brock2018large} (with trunc.)                        & 112  & 1   & 7.03  & 0.872 & 0.277 \\
StyleGAN-XL\cite{stylegan-xl} (without trunc.)                       & 166  & 1   & 6.284  & 0.919 & 0.261 \\
StyleGAN-XL\cite{stylegan-xl} (with trunc.)                      & 166  & 1   & 2.45  & 0.812 & 0.490 \\
\midrule
DiT-L/2~\cite{peebles2023scalable}                  & 458  & 250 & 23.3  & {-}   & {-}   \\
DiT-XL/2                             & 675  & 250 & 9.6  & {-} & {-} \\
DiT-XL/2 (cfg $= 1.5$)                              & 675  & 250 & 2.27  & 0.830 & 0.57 \\
\midrule
SiT-L/2~\cite{ma2024sit}                            & 458  & 250 & 18.8  & {-}   & {-}   \\
SiT-XL/2                               & 675  & 250 & 8.3  & {-} & {-} \\
SiT-XL/2 (cfg $= 1.5$)                              & 675  & 250 & 2.06  & 0.830 & 0.593 \\
\midrule
iMeanFlow \cite{geng2025improved}                    & 610 & 1   & 1.72  & 0.78150 & 0.62570 \\
\midrule
IMM (1 step, cfg $= 1.5$) \cite{zhou2025inductive}   & 675  & 1   & 8.05  & 0.66624  &  0.64330 \\
IMM (2 steps, cfg $= 1.5$)                          & 675  & 2   & 3.99  & 0.73572  &  0.64610 \\
\midrule
VAR-$d$16 \cite{tian2024visualautoregressivemodelingscalable}     & 310  & 10  & 3.30  & 0.84  & 0.51  \\
VAR-$d$30-re                          & 2000 & 10  & 1.73  & 0.82  & 0.60  \\
\midrule
\textit{\methodshort} (Latent)                                                & 310  & 1   & 4.16  & 0.77  & 0.49  \\
\textit{\methodshort} (Latent) + Round-trip rejection                                                & 310  & 1   & 2.56  & 0.79  & 0.50  \\
\bottomrule
\end{tabular}%
}
\end{table}
\subsection{Results and Analysis} 

\textbf{CIFAR-10} Table~\ref{tab:cifar10-results} shows that our model attains the best precision and recall in the table by wide margins: precision of 0.91 against 0.70 for the next best method, and recall of 0.80 against 0.60. Notably, these margins hold even against the 100-step DDIM baseline (0.62 / 0.57), so a single forward pass of our model covers the data distribution better than a hundred evaluations of a diffusion sampler. At one NFE (number of function evaluations), our FID is competitive with the strongest one-step baselines.

\textbf{CelebA-HQ} Table~\ref{tab:celeba} shows that our model has the best precision and recall, while being the second-best on FID. The 128-step Shortcut model~\cite{frans2024stepdiffusionshortcutmodels} is outperformed on every metric despite it using more evaluation steps. Compared to StyleSwin, which achieves the best FID, our model produces more diverse samples (recall 0.61 vs.\ 0.39).

\textbf{ImageNet 256} Table~\ref{tab:imagenet-results} shows that our latent-space model reaches an FID of 4.16 at one NFE with 310M parameters. With round-trip rejection, our FID improves to 2.56. 
Our FID is competitive with strong one-NFE and multi-step baselines such as iMeanFlow, StyleGAN-XL, DiT-XL/2, and SiT-XL/2.

\begin{figure*}
  \captionsetup[subfigure]{labelformat=parens}

    \includegraphics[width=\linewidth]{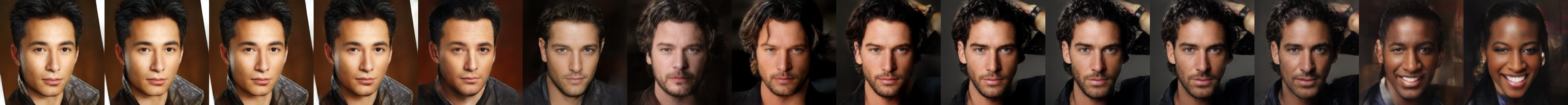}
    \includegraphics[width=\linewidth]{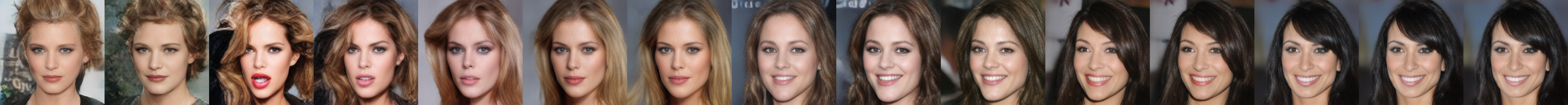}

  \caption{\textbf{Prior-space interpolation for CelebA-HQ}. Interpolating linearly between two prior samples $\mathbf{z}$ and decoding each intermediate point produces plausible images that vary smoothly, suggesting the generator has learned a coherent mapping from the prior rather than memorizing the training data.}
\end{figure*}

\begin{figure*}
  \begin{center}
    \includegraphics[width=0.9\linewidth]{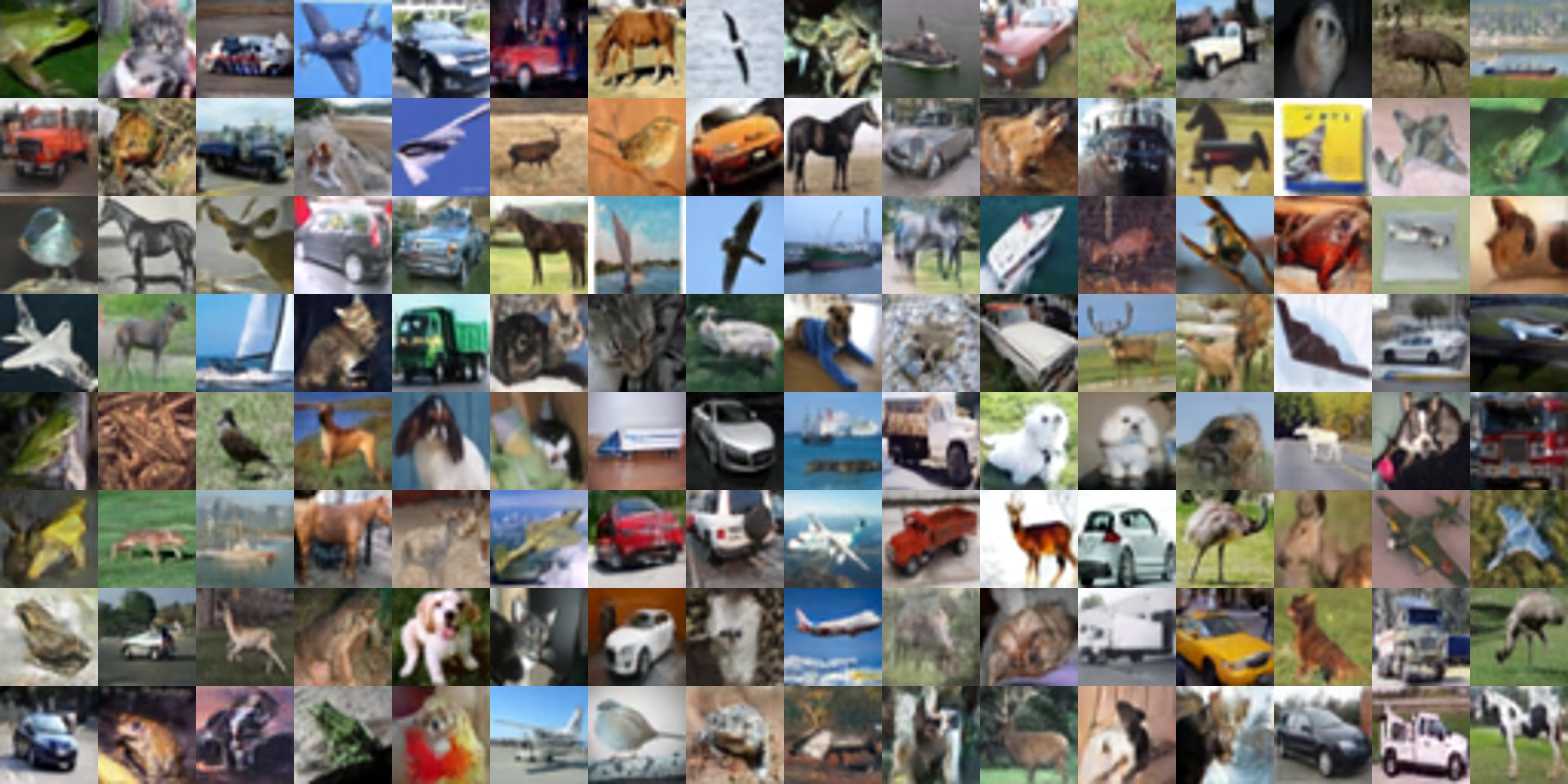}
  \end{center}
  \caption{Random samples from our unconditional model trained on CIFAR-10}

  \label{fig:cifar10-samples}
\end{figure*}

\section{Discussion and Conclusion}

% \textbf{Limitations} Although \methodshort{} substantially outperforms vanilla IMLE, it processes images across multiple resolutions during training, which marginally increases training time.

\textbf{Societal Impact}
We are aware that generative models can facilitate creating fake images (``deepfakes'') that can be exploited for disinformation or fraudulent activities, but note that watermarking techniques, e.g.,~\cite{gunn2024undetectable}, have been developed that can mitigate such harms. 

\textbf{Conclusion}
We introduced \methodshort{}, a single-step IMLE-based generative model. The method modifies IMLE training in two ways: multi-stage supervision at each upsampling resolution, and a robust loss that handles mismatched nearest-neighbour pairs. Across CIFAR-10, CelebA-HQ, and ImageNet 256, these changes make \methodshort{} competitive in FID while attaining strong precision and recall among one-step methods. More broadly, our results show that a single-step generator like IMLE can match much of the quality and diversity of iterative models. 

\textbf{Acknowledgements}
This research was enabled in part by NSERC, the Canada Foundation for Innovation, the Canada CIFAR AI Chairs program, the BC DRI Group and the Digital Research Alliance of Canada.

\clearpage

\bibliographystyle{plainnat}  % or abbrvnat, unsrtnat, etc.
\bibliography{main}   
%%%%%%%%%%%%%%%%%%%%%%%%%%%%%%%%%%%%%%%%%%%%%%%%%%%%%%%%%%%%

\clearpage
\appendix

\section{Note on Maximum Likelihood Estimation}
\label{appendix-mle}

Maximum Likelihood Estimation (MLE) is widely used in generative modeling because it provides a principled method for fitting models to data by maximizing the likelihood that observed samples originate from the model's distribution. Consequently, samples generated from models optimized with MLE (and its variants) are, by design, expected to closely resemble the data points. Directly optimizing the likelihood is often computationally intractable, since the data is generally high-dimensional and the model's mapping from the prior to the data distribution is typically not surjective. Different generative models take different approaches to get around this issue.

Diffusion models, for example, circumvent this by maximizing a variational lower bound known as the Evidence Lower Bound (ELBO). Specifically, diffusion models factorize the generation process into multiple small, iterative transitions from a known prior distribution $p(\mathbf{x}_T)$. This can be written as follows:
\begin{align}
    p(\mathbf{x}_0 \mid \mathbf{x}_{T}) &=  \int \cdots \int  p_\theta(\mathbf{x}_{0} \mid \mathbf{x}_{1}) \,
    p_\theta(\mathbf{x}_{1} \mid \mathbf{x}_{2}) \cdots
    p_\theta(\mathbf{x}_{T-1} \mid \mathbf{x}_{T})
    \,\mathrm{d}\mathbf{x}_1 \cdots \mathrm{d}\mathbf{x}_{T-1} \notag \\
    &= \int \cdots \int \left[ \prod_{t=1}^{T} p_\theta(\mathbf{x}_{t-1} \mid \mathbf{x}_t) \right] \, \mathrm{d}\mathbf{x}_1 \cdots \mathrm{d}\mathbf{x}_{T-1}
\end{align}
In contrast, Implicit Maximum Likelihood Estimation (IMLE) models the transition from the prior to the data in a single step, implicitly optimizing the MLE without explicitly factorizing into intermediate distributions.

\section{Experiment details for ablation study on Oxford Flowers}
\label{appendix-ablation-flowers}

This section provides additional details for the ablation study on the Oxford Flowers dataset, presented in Table~\ref{tab:method-ablation} of the main paper. 
All configurations (A through E) use the identical architecture and the exact training setup; the only variable changed is the training objective. The only minor difference is that multi-resolution configurations have a trainable $1\times1$ convolution layer for intermediate outputs per stage. This change slightly increases the number of parameters for configurations B-E by $\text{network width} \times \text{output channels} \times \text{number of stages}$. In practice, it is close to roughly 4K extra parameters (varies slightly depending on the dataset), on the order $10^{-5}$ of the model size. In our actual implementation, we just set these parameters to not be trainable for vanilla IMLE (configuration A).

All the configurations were trained for exactly the same number of steps. Comparing method variants for the same number of training iterations (epochs) under an identical training setting is standard practice in the literature~\cite{eqvae, stoica2025contrastiveflowmatching}.

\begin{table}[h]
\centering
\caption{FID $\downarrow$ over training for the five configurations of Table~\ref{tab:method-ablation}, reported as mean $\pm$ standard deviation over 3 seeds.}
\label{tab:ablation-convergence}
\small
\resizebox{\textwidth}{!}{%
\begin{tabular}{lcccccccc}
\toprule
& \multicolumn{8}{c}{Epoch} \\
\cmidrule(lr){2-9}
Conf. & 500 & 1000 & 1500 & 2000 & 2500 & 3000 & 3500 & 3900 \\
\midrule
A & 169.66 $\pm$ 4.17 & 134.31 $\pm$ 5.30 & 115.20 $\pm$ 3.65 & 101.40 $\pm$ 2.44 & 88.91 $\pm$ 4.48 & 75.71 $\pm$ 1.71 & 68.03 $\pm$ 1.40 & 62.61 $\pm$ 2.28 \\
\midrule
B & 142.41 $\pm$ 2.25 & 105.56 $\pm$ 6.23 & 82.57 $\pm$ 3.72 & 65.16 $\pm$ 3.29 & 51.86 $\pm$ 2.17 & 43.66 $\pm$ 1.93 & 34.77 $\pm$ 1.06 & 31.24 $\pm$ 1.08 \\
\midrule
C & 108.94 $\pm$ 5.00 & 67.56 $\pm$ 2.92 & 51.12 $\pm$ 9.02 & 40.74 $\pm$ 1.13 & 30.10 $\pm$ 1.73 & 25.12 $\pm$ 1.14 & 21.30 $\pm$ 0.66 & 19.80 $\pm$ 0.58 \\
D & 97.02 $\pm$ 4.45 & 55.79 $\pm$ 4.53 & 36.96 $\pm$ 1.45 & 28.84 $\pm$ 0.63 & 24.01 $\pm$ 1.61 & 20.59 $\pm$ 1.61 & 16.44 $\pm$ 0.58 & 16.49 $\pm$ 0.09 \\
E & \textbf{83.52 $\pm$ 5.43} & \textbf{46.26 $\pm$ 3.55} & \textbf{35.24 $\pm$ 2.32} & \textbf{24.57 $\pm$ 1.45} & \textbf{18.77 $\pm$ 0.17} & \textbf{15.97 $\pm$ 0.91} & \textbf{15.42 $\pm$ 0.60} & \textbf{13.90 $\pm$ 0.08} \\
\bottomrule
\end{tabular}%
}
\end{table}

\begin{figure}[h]
\centering
\includegraphics[width=0.9\linewidth]{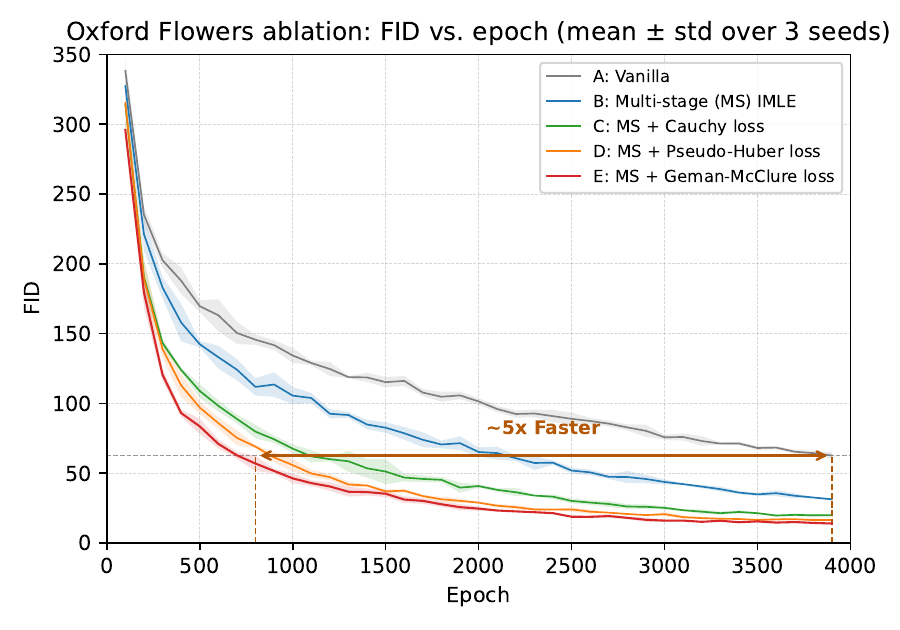}
\caption{FID vs.\ epoch on Oxford Flowers for the five configurations of Table~\ref{tab:method-ablation}, reported as mean $\pm$ standard deviation over 3 seeds. Our proposed configuration (E) reaches the baseline's (A) final FID roughly 5x faster.}
\label{fig:ablation-convergence}
\end{figure}

Our proposed method converges roughly $5 \times$ faster than the original under identical training conditions: it reaches the baseline's final FID (62.61 at 3900 epochs) in $\sim 800$ epochs. Crucially, at every matched step count our proposed method's FID is strictly lower. Table~\ref{tab:ablation-convergence} reports FID at 500-epoch intervals for all five configurations, each as mean $\pm$ standard deviation over 3 seeds, and Figure~\ref{fig:ablation-convergence} plots the full FID trajectory over training.

\section{Ablation on Sampling Budget}
\label{appendix-sampling-budget}

The IMLE matching phase selects nearest neighbours for $n$ data points among $m$ generated samples. In this section, we ablate the sampling budget $m/n$. For our configuration E from Table~\ref{tab:method-ablation}, we report the effect of the sampling budget on FID for Oxford-Flowers. We vary the ratio $m/n$ and report FID at different training epochs in Table~\ref{tab:sampling-budget}. Even with a small sample factor (2), our method converges and reaches reasonable performance. The gains show clear diminishing returns: moving from 4 to 10 improves the final FID only marginally. These results show that IMLE does not need to generate a large number of samples in the matching phase to achieve good performance.

\begin{table}[h]
\centering
\caption{Effect of the nearest-neighbour sampling budget on FID for Oxford-Flowers.}
\label{tab:sampling-budget}
\small
\begin{tabular}{lccc}
\toprule
\textbf{Epoch} & $m/n = 2$ & $m/n = 4$ & $m/n = 10$ \\
\midrule
500  & 94.42 & 78.65 & 66.37 \\
1000 & 64.60 & 49.38 & 39.54 \\
1500 & 46.04 & 35.92 & 26.76 \\
2000 & 36.04 & 25.80 & 20.59 \\
2500 & 26.58 & 18.88 & 17.68 \\
3000 & 27.34 & 16.79 & 15.00 \\
3500 & 23.56 & 14.85 & 13.41 \\
\bottomrule
\end{tabular}
\end{table}

For the experiments in Section~\ref{experiments}, we use a budget of $m/n = 5$ across all three datasets (CIFAR-10, CelebA-HQ, ImageNet 256).
\section{Architecture and training details}
\label{appendix-details}

\begin{figure}[ht]
    \centering
    \subfloat[Network Architecture]{
    \includegraphics[width=0.53\linewidth]{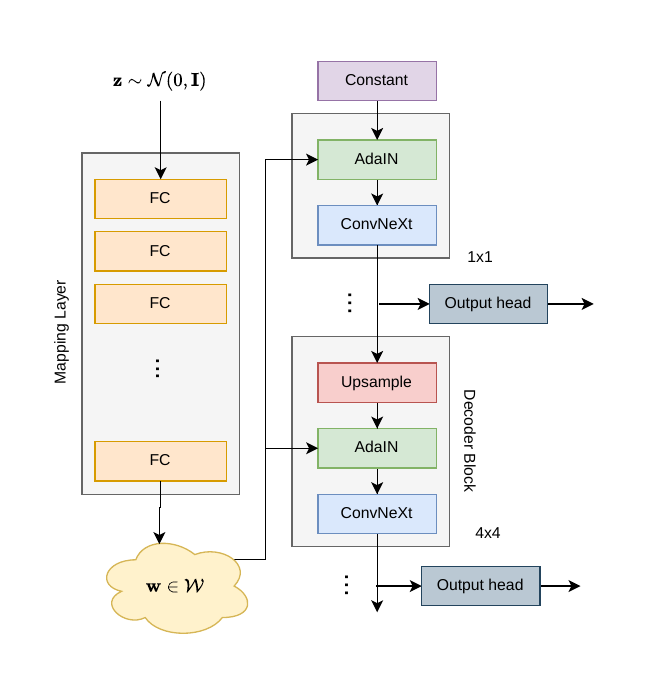}
    \label{fig:overall}
  }
  \subfloat[Standard ConvNeXt block]{
    \includegraphics[width=0.3\linewidth]{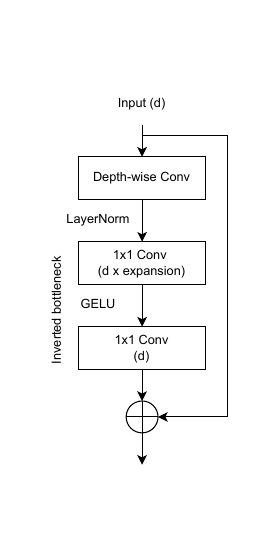}
    \label{fig:res}
  }
    \caption{(a) Network architecture, which comprises a mapping network, upsampling layers and ConvNeXt blocks. (b) Inner workings of a standard ConvNeXt block.}
    \label{fig:arch}
\end{figure}

We provide the detailed network architecture in Figure~\ref{fig:arch}.\\
Our generator maps a latent code ($\mathbf{z} \sim \mathcal{N}(0,\mathbf{I}), \mathbf{z} \in \mathbb{R}^{1024}$) to a style vector $\mathbf{w}$ via an MLP (mapping layer), which modulates each layer of a ConvNeXt-style decoder using Adaptive Instance Normalization (AdaIN)\cite{adain}. Starting from a learned constant feature map at low resolution, the decoder progressively upsamples the representation through a stack of residual ConvNeXt\cite{liu2022convnet2020s} blocks with per-resolution width control.

\textbf{Training Details}
The training was done on NVIDIA L40S GPUs. The training time depends primarily on the size of the dataset. Default PyTorch implementation of bicubic interpolation was used for all resizing operations, because it achieved the best results. We used a cosine scheduler with warmup for 100 iterations. 

\textbf{Pixel-based models} We used a weighted combination of LPIPS \cite{lpips} loss, DINO \cite{dino} loss and pixel loss for optimization, with weighting factors of 1.0, 1.0 and 0.1 respectively. Since DINO expects $224\times224$ inputs, all images were resized before being passed into the DINO network. LPIPS requires images to be of at least $32\times32$ resolution; hence images below this resolution were resized to $32\times32$ before being passed into the LPIPS network.
The $32\times32$ variant (used for CIFAR-10) contains 110 million trainable parameters, while the $256\times256$ variant (used for CelebA-HQ) has 138 million. We do not use any data augmentation.

\textbf{Latent-based models} For ImageNet, we train our generator in the latent space of an EQ-VAE \cite{eqvae}, which is kept frozen throughout training. The model contains 310 million trainable parameters. The per-stage training loss is the Geman-McClure loss applied to elementwise residuals computed directly on the latent representations. Per-stage targets are obtained by encoding the real image with the frozen EQ-VAE encoder and downsampling the resulting latent map with $K_l$ to each stage's resolution.

We use rejection sampling to remove low-quality samples, with a different rejection criterion: at evaluation time, we compute for each generated sample the round-trip reconstruction cost of encoding and decoding it with the EQ-VAE. The distance between the generated sample and its round-trip reconstruction is measured in LPIPS \cite{lpips} space.
Low-quality samples incur a high round-trip cost, and we reject samples whose cost exceeds a fixed threshold, which removes roughly 5\% of generated samples. We find that removing these samples improves FID. Reported ImageNet metrics are computed on samples that pass this filter.

We provide additional details about the training setup that are relevant for reproduction and for clarifying the experimental protocol.

\textbf{Distance metric for nearest-neighbour selection.} The distance $d(\cdot, \cdot)$ used in the $\arg\min_j$ step of Algorithm~\ref{alg:our_proc} differs between pixel-space and latent-space models. For our pixel-space models (CIFAR-10, CelebA-HQ), we use LPIPS~\cite{lpips} as the selection distance. For our latent-space model on ImageNet 256, we use the squared $\ell_2$ distance in the EQ-VAE~\cite{eqvae} latent space.

\textbf{Nearest-neighbour search at scale.} We use FAISS-GPU to accelerate exact nearest-neighbour search across the $m$ candidate samples.

\section{Note on Evaluation Metrics} 
\label{appendix-eval}

In order to evaluate a generative model, we need a holistic picture across complementary metrics. While FID remains the most widely reported, it is known to have serious flaws: it compresses everything into a single scalar and cannot indicate what a model is lacking. In particular, \citet{rethinking-fid} shows that two very different distributions can be considered identical under the Gaussian feature assumption of the Fréchet distance. Hence metrics like precision and recall, which separately measure fidelity (how realistic samples are) and coverage (how well the data distribution is covered), give a more complete assessment. For our case, high precision and recall indicate that individual samples are realistic and that the generated distribution covers the real data well: the two properties one ultimately cares about in a generative model. We read the combination of strong precision/recall with competitive FID as evidence that the method produces realistic, diverse samples, rather than optimizing a single distributional statistic.

For FID, we calculate the Fréchet distance between the real and generated distributions in the feature space of a pre-trained InceptionV3 network using 50K generated images. We report precision and recall using the standard implementation of \citet{kynkaanniemi2019improved}, with the InceptionV3 feature extractor and $k = 3$ for the $k$-nearest-neighbour radius. This matches the protocol most commonly used in prior image generation literature. Not all baselines report precision and recall, so we computed the missing numbers ourselves. Protocols also vary across papers, most notably in the Inception weights that the implementation uses, and these choices can shift the resulting values. To have consistency, we use the most popular protocol described by \citet{diffusion-beats-gan} and apply it uniformly to every number we computed.

\section{Nearest-neighbour test}
\label{appendix-nntest}

\begin{figure*}[b!]
  \centering
  \small
  \begin{tabular}{@{}>{\centering\arraybackslash}m{0.10875\linewidth}@{\hspace{8pt}}>{\centering\arraybackslash}m{0.87\linewidth}@{}}
    \textcolor{orange}{\textbf{Generated}} & \textcolor{blue}{\textbf{Nearest neighbours from training set}} \\[6pt]
    \begin{tikzpicture}
      \node[inner sep=0] (img) {\includegraphics[width=\linewidth]{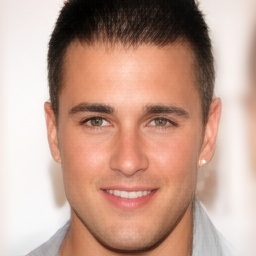}};
      \draw[orange, line width=1pt] (img.south west) rectangle (img.north east);
    \end{tikzpicture}
    &
    \begin{tikzpicture}
      \node[inner sep=0] (img) {\includegraphics[width=\linewidth]{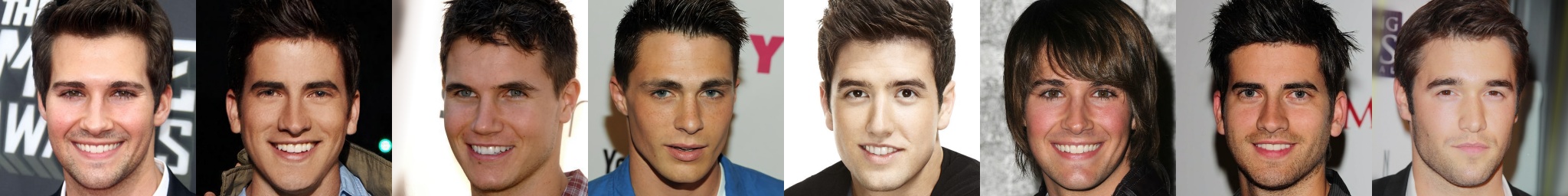}};
      \draw[blue, line width=1pt] (img.south west) rectangle (img.north east);
    \end{tikzpicture} \\[4pt]
    \begin{tikzpicture}
      \node[inner sep=0] (img) {\includegraphics[width=\linewidth]{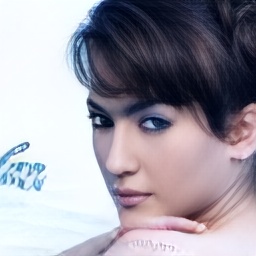}};
      \draw[orange, line width=1pt] (img.south west) rectangle (img.north east);
    \end{tikzpicture}
    &
    \begin{tikzpicture}
      \node[inner sep=0] (img) {\includegraphics[width=\linewidth]{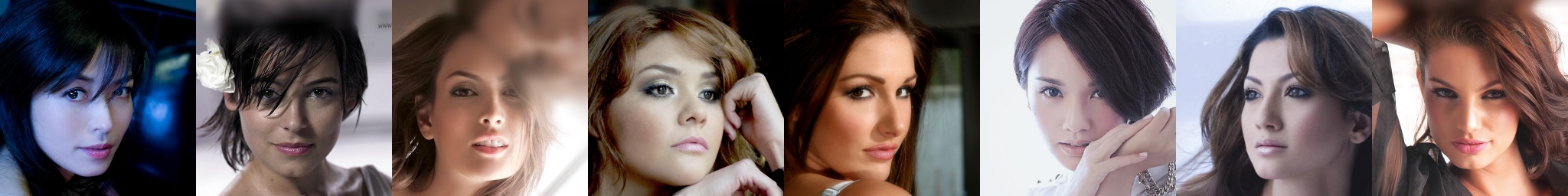}};
      \draw[blue, line width=1pt] (img.south west) rectangle (img.north east);
    \end{tikzpicture} \\[4pt]
    \begin{tikzpicture}
      \node[inner sep=0] (img) {\includegraphics[width=\linewidth]{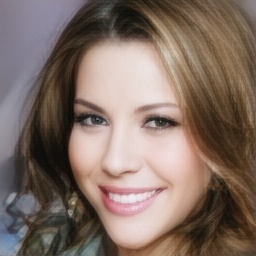}};
      \draw[orange, line width=1pt] (img.south west) rectangle (img.north east);
    \end{tikzpicture}
    &
    \begin{tikzpicture}
      \node[inner sep=0] (img) {\includegraphics[width=\linewidth]{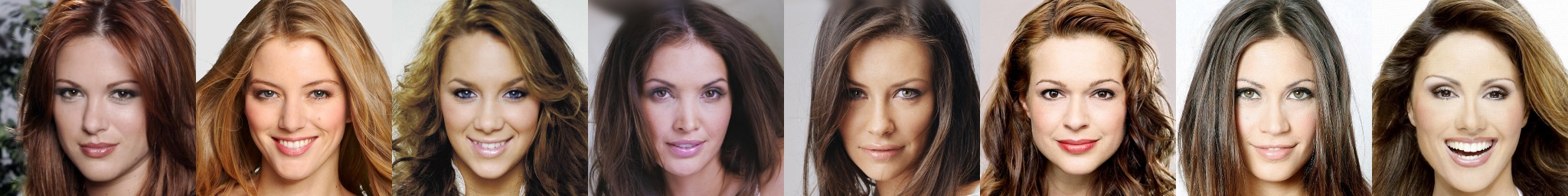}};
      \draw[blue, line width=1pt] (img.south west) rectangle (img.north east);
    \end{tikzpicture} \\[4pt]
    \begin{tikzpicture}
      \node[inner sep=0] (img) {\includegraphics[width=\linewidth]{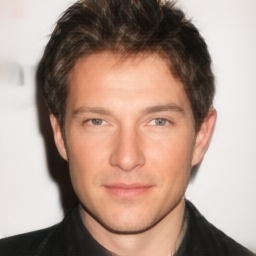}};
      \draw[orange, line width=1pt] (img.south west) rectangle (img.north east);
    \end{tikzpicture}
    &
    \begin{tikzpicture}
      \node[inner sep=0] (img) {\includegraphics[width=\linewidth]{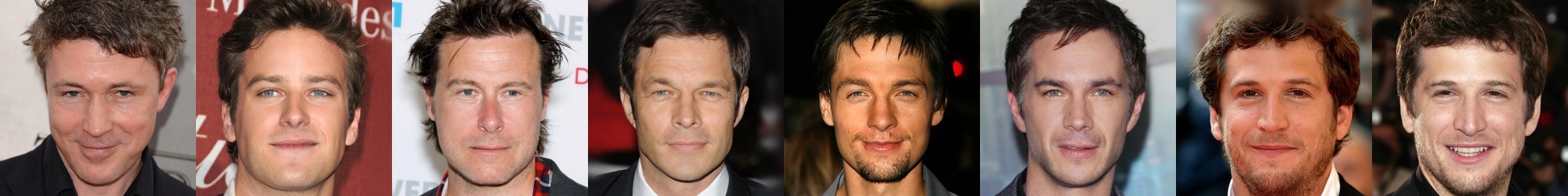}};
      \draw[blue, line width=1pt] (img.south west) rectangle (img.north east);
    \end{tikzpicture} \\[4pt]
    \begin{tikzpicture}
      \node[inner sep=0] (img) {\includegraphics[width=\linewidth]{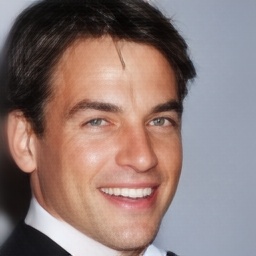}};
      \draw[orange, line width=1pt] (img.south west) rectangle (img.north east);
    \end{tikzpicture}
    &
    \begin{tikzpicture}
      \node[inner sep=0] (img) {\includegraphics[width=\linewidth]{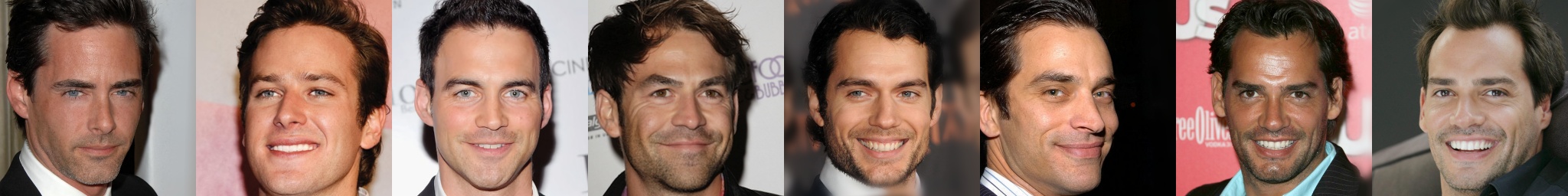}};
      \draw[blue, line width=1pt] (img.south west) rectangle (img.north east);
    \end{tikzpicture} \\[4pt]
    \begin{tikzpicture}
      \node[inner sep=0] (img) {\includegraphics[width=\linewidth]{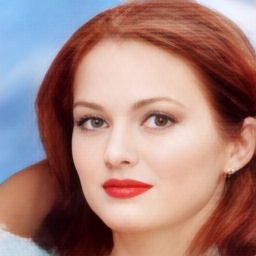}};
      \draw[orange, line width=1pt] (img.south west) rectangle (img.north east);
    \end{tikzpicture}
    &
    \begin{tikzpicture}
      \node[inner sep=0] (img) {\includegraphics[width=\linewidth]{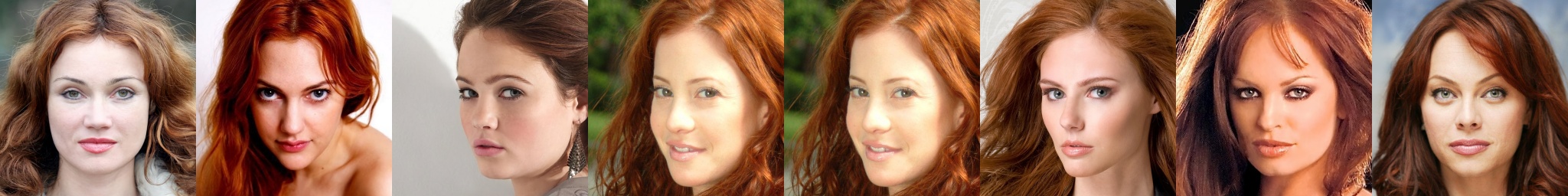}};
      \draw[blue, line width=1pt] (img.south west) rectangle (img.north east);
    \end{tikzpicture} \\[4pt]
    \begin{tikzpicture}
      \node[inner sep=0] (img) {\includegraphics[width=\linewidth]{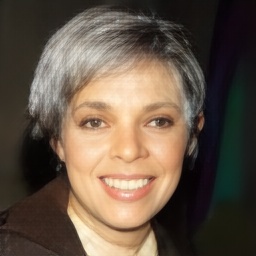}};
      \draw[orange, line width=1pt] (img.south west) rectangle (img.north east);
    \end{tikzpicture}
    &
    \begin{tikzpicture}
      \node[inner sep=0] (img) {\includegraphics[width=\linewidth]{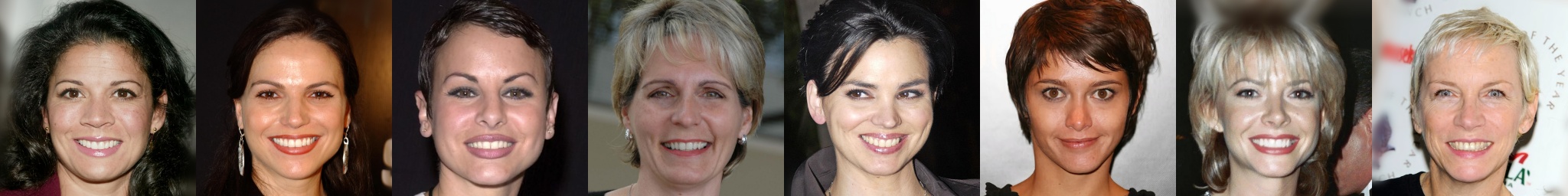}};
      \draw[blue, line width=1pt] (img.south west) rectangle (img.north east);
    \end{tikzpicture} \\[4pt]
    \begin{tikzpicture}
      \node[inner sep=0] (img) {\includegraphics[width=\linewidth]{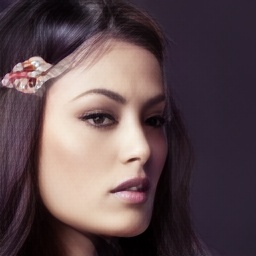}};
      \draw[orange, line width=1pt] (img.south west) rectangle (img.north east);
    \end{tikzpicture}
    &
    \begin{tikzpicture}
      \node[inner sep=0] (img) {\includegraphics[width=\linewidth]{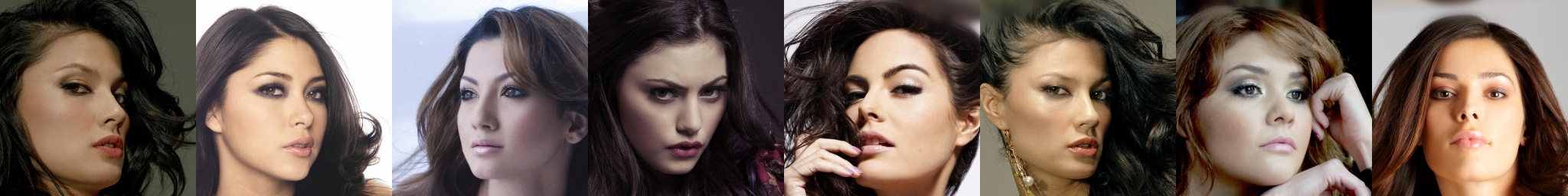}};
      \draw[blue, line width=1pt] (img.south west) rectangle (img.north east);
    \end{tikzpicture} \\[4pt]
    \begin{tikzpicture}
      \node[inner sep=0] (img) {\includegraphics[width=\linewidth]{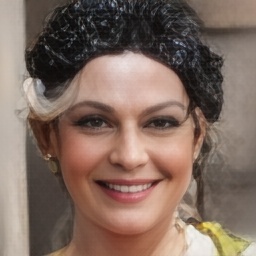}};
      \draw[orange, line width=1pt] (img.south west) rectangle (img.north east);
    \end{tikzpicture}
    &
    \begin{tikzpicture}
      \node[inner sep=0] (img) {\includegraphics[width=\linewidth]{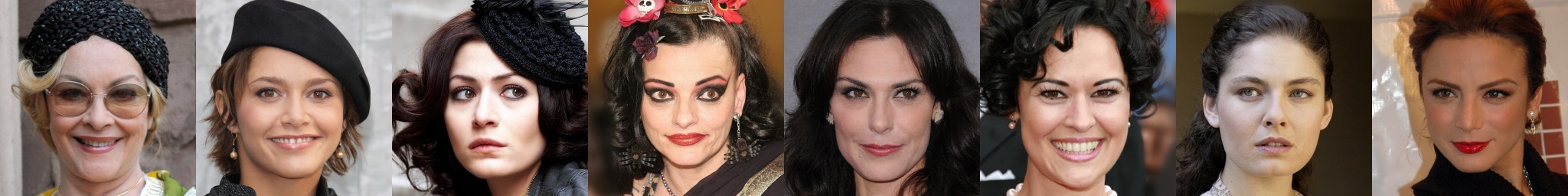}};
      \draw[blue, line width=1pt] (img.south west) rectangle (img.north east);
    \end{tikzpicture} \\[4pt]
  \end{tabular}
  \caption{\textbf{Nearest-neighbour analysis on CelebA-HQ.} Each row shows a \textcolor{orange}{generated sample} together with its \textcolor{blue}{nearest neighbours from the training set}, retrieved by cosine similarity using CLIP encoder~\cite{clip}. Our generated samples are visually distinct from their nearest neighbours, indicating that the model is not memorizing training images.}
  \label{fig:nn-test}
\end{figure*}
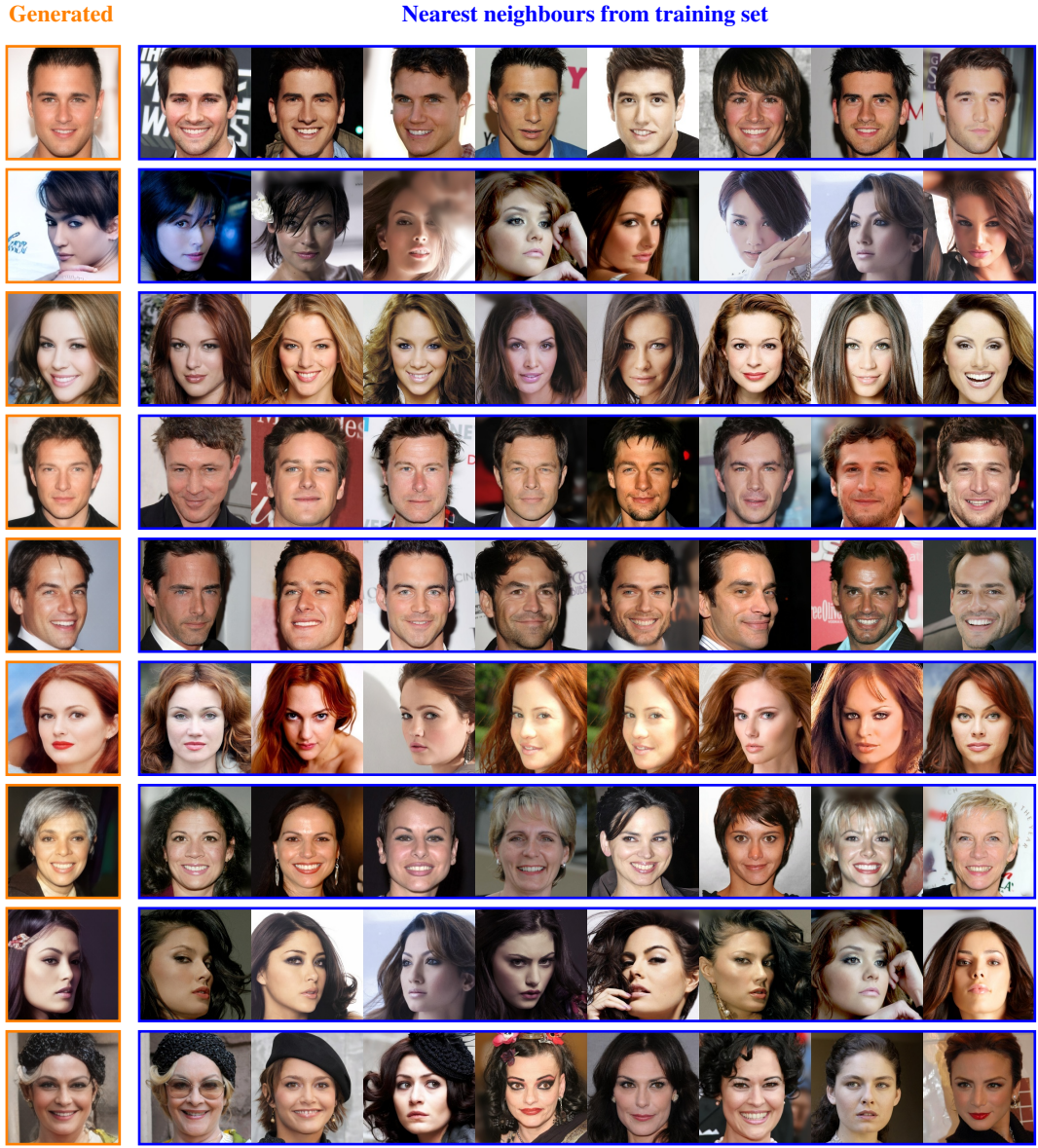

A natural concern for a generative model trained with a nearest-neighbour matching objective is that the generator could memorize training images rather than synthesize novel ones. To check for this, we retrieve the top-8 nearest neighbours of each generated sample from the CelebA-HQ training set, using cosine similarity in CLIP embedding space~\cite{clip}. Figure~\ref{fig:nn-test} shows this comparison for a set of generated samples. The results indicate that our model synthesizes novel images rather than reproducing individual training examples.

% \newpage
% \input{checklist.tex}

\end{document}